\documentclass[mathpazo]{cicp}

\usepackage{amsmath,amsfonts,bm}

\def\eqref#1{equation~\ref{#1}}

\def\1{\bm{1}}

\def\vu{{\bm{u}}}

\def\vx{{\bm{x}}}

\DeclareMathAlphabet{\mathsfit}{\encodingdefault}{\sfdefault}{m}{sl}
\SetMathAlphabet{\mathsfit}{bold}{\encodingdefault}{\sfdefault}{bx}{n}

\usepackage{lipsum}
\usepackage{amsfonts}
\usepackage{epstopdf}
\usepackage{algorithmic}

\usepackage{xcolor}
\usepackage{hyperref}
\usepackage{cleveref}

\crefname{figure}{Figure}{Figure}
\crefname{table}{Table}{Table}
\crefname{equation}{Eq.}{Eq.}
\crefname{appendix}{Appendix}{Appendix}
\crefname{section}{Section}{Section}
\usepackage{url}
\usepackage{booktabs}
\usepackage{graphics}
\usepackage{bm}
\usepackage{graphicx}
\usepackage{subfigure}
\usepackage{amsmath}
\usepackage{colortbl}
\usepackage{bbding}

\begin{document}
\title{Learning Specialized Activation Functions for Physics-informed Neural Networks}


\author[Wang H H et.~al.]{Honghui Wang\affil{1},
      Lu Lu\affil{2}, Shiji Song\affil{1} and Gao Huang\affil{1}\comma\corrauth}
\address{\affilnum{1}\ Department of Automation,
         Tsinghua University,
         Beijing 100084, P.R. China. \\
          \affilnum{2}\ Department of Chemical and Biomolecular Engineering,
          University of Pennsylvania, Philadelphia, PA 19104, USA}
\emails{{\tt wanghh20@mails.tsinghua.edu.cn} (H.~Wang), {\tt lulu1@seas.upenn.edu} (L.~Lu),
         {\tt shijis@mail.tsinghua.edu.cn} (S.~Song), {\tt gaohuang@tsinghua.edu.cn} (G.~Huang)}

\begin{abstract}
Physics-informed neural networks (PINNs) are known to suffer from optimization difficulty. In this work, we reveal the connection between the optimization difficulty of PINNs and activation functions. Specifically, we show that PINNs exhibit high sensitivity to activation functions when solving PDEs with distinct properties. Existing works usually choose activation functions by inefficient trial-and-error. To avoid the inefficient manual selection and to alleviate the optimization difficulty of PINNs, we introduce adaptive activation functions to search for the optimal function when solving different problems. We compare different adaptive activation functions and discuss their limitations in the context of PINNs. Furthermore, we propose to tailor the idea of learning combinations of candidate activation functions to the PINNs optimization, which has a higher requirement for the smoothness and diversity on learned functions. This is achieved by removing activation functions which cannot provide higher-order derivatives from the candidate set and incorporating elementary functions with different properties according to our prior knowledge about the PDE at hand. We further enhance the search space with adaptive slopes. The proposed adaptive activation function can be used to solve different PDE systems in an interpretable way. Its effectiveness is demonstrated on a series of benchmarks. Code is available at https://github.com/LeapLabTHU/AdaAFforPINNs.
\end{abstract}


\maketitle

\newcommand{\name}{ABU-PINN}

\section{Introduction}
 
Recent years have witnessed the remarkable progress of physics-informed neural networks (PINNs) on modeling the dynamics of physics systems~\cite{raissi2019physics,chen2020physics,shin2020convergence,lu2021physics,karniadakis2021physics,jiao2022rate}.
The underlying physics laws, usually presented as ordinary and partial differential equations (ODEs and PDEs), are embedded as soft constraints to guide the learning process of deep neural networks~\cite{weinan2018deep,raissi2019physics,zhang2021mod}.
Despite the effectiveness, the introduction of PDE-based loss function makes the optimization more ill-conditioned~\cite{wang2021understanding,krishnapriyan2021characterizing}. Efforts have been made to alleviate this problem from the aspects of loss weight balancing~\cite{wang2021understanding,wang2022and}, loss function design~\cite{psaros2022meta,yu2022gradient,wang20222}, adaptive collocation point sampling~\cite{zhao2020solving,mcclenny2020self,lu2021deepxde,wu2022comprehensive}, domain decomposition~\cite{jagtap2020conservative,jagtap2020extended,moseley2021finite}, and curriculum learning~\cite{krishnapriyan2021characterizing}. However, since most previous works adopt the standard fully-connected networks, the relationship between network architectures and the optimization difficulty of PINNs is less explored, regardless of the fact that the advancement of network architectures is the key to the success on deep learning in computer vision~\cite{he2016deep,huang2017densely} and natural language processing~\cite{vaswani2017attention}.   

At the heart of network architectures lie the activation functions, which play a significant role in the expressiveness and optimization of models. Recent works~\cite{raissi2019deep,sitzmann2020implicit,xu2020frequency,CiCP-28-1886,wong2022learning} have observed that the choice of activation functions affects the learning of continuous signal representations. 
For example, the hyperbolic tangent function is shown to suffer from numerical instability when simulating vortex induced vibrations, while a PINN with sinusoidal function can be optimized smoothly~\cite{raissi2019deep}. 
Another important observation is that the optimal activation function depends on the problem at hand. While the Rectified Linear Unit~\cite{hahnloser2000digital,jarrett2009best,nair2010rectified,lu2020dyingrelu} is widely adopted in most computer vision and natural language processing tasks~\cite{ramachandran2017searching}, there is no such default choice of activation functions for PINNs when applied to physical systems with distinct properties. In fact, PINNs show great sensitivity to activation functions.
These observations reveal the possibility and necessity of reducing the training difficulty of PINNs by selecting an appropriate activation function. 

The various characteristics of different PDE systems make the choice of activation functions a critical aspect in PINNs. The common practice  to find the optimal activation functions is by trial-and-error, which requires extensive computational resources and human knowledge. 
As an alternative, the adaptive activation functions are designed to find specialized activation functions for different architectures and tasks. In this work, we aim to explore different adaptive activation functions and discuss their limitations in the context of PINNs. The major difference of existing methods lies in the search space. The piece-wise linear function is adopted as the universal function approximator in some methods, such as APL~\cite{agostinelli2014learning}, RePLU~\cite{li2016sparseness} and PWLU~\cite{zhou2021learning}. The formulation of SLAF~\cite{goyal2019learning} is based on Taylor approximation with polynomial basis. PAU~\cite{molina2019pad} leverages Pad\'e approximation to form its search space.  
ACON~\cite{ma2021activate} is proposed as a smooth approximator to the general Maxout family activation functions~\cite{goodfellow2013maxout}. 
Built upon previous works~\cite{dushkoff2016adaptive,qian2018adaptive,manessi2018learning} to learn activation functions as linear combinations of candidate activation functions, the Adaptive Blending Units (ABU)~\cite{sutfeld2020adaptive} explore different normalization methods for the combination coefficients, which correspond to different restrictions imposed on the search space. 
In the literature of PINNs, the recent work~\cite{jagtap2022deep} proposes the Kronecker neural network (KNN) based on adaptive linear combinations of candidate activation functions and apples it to solve PDEs. However, only one type of adaptive activation function is discussed in their work under the context of PINNs, and the designs of its candidate function set and normalization method of combination coefficients are yet to be explored to better suit the optimization of PINNs. 

Since most existing works mainly focus on convolutional neural networks for the image classification task, obstacles exist in their direct application in the context of PINNs, which have a higher demand for the smoothness and diversity of the learned functions. First, the optimization of PDE-based constraints needs the activation function to provide higher-order derivatives, which causes the failure of widely-used ReLUs and other piecewise linear functions~\cite{agostinelli2014learning,li2016sparseness,zhou2021learning} in PINNs. Second, unlike the image classification tasks, different PDE systems could have distinct characteristics, such as periodicity and rapid decay. This leads to a higher requirement for the diversity of the learned functions.

To avoid the inefficient manual selection of activation functions and to alleviate the optimization difficulty of PINNs, we propose the Adaptive Blending Units for PINNs (ABU-PINN) to combine the simple idea of learning combinations of candidate activation functions with special treatments tailored for PINN optimizations. We first remove those activation functions which cannot provide continuous and non-zero higher-order derivatives from the candidate function set. To deal with the various characteristics of PDEs, we incorporate elementary functions with different properties as candidate functions according to our prior knowledge of a physics system at hand. 
For instance, we can leverage the sinusoidal and exponential functions to reduce the modeling difficulty of periodicity and rapid decay phenomena, which are commonly found in physical systems. 
Note that we reveals that adaptive activation functions can be more powerful and interpretable in handling PDE systems than they are in general domains like image classification or natural language processing~\cite{dushkoff2016adaptive,qian2018adaptive,manessi2018learning,sutfeld2020adaptive}, because there is no such prior knowledge to leverage. Besides these elementary functions, we also add most of commonly-used activation functions to ensure the diversity of the candidate set. Finally, we further enhance the search space with adaptive slopes~\cite{jagtap2020adaptive,jagtap2020locally}. 
The adaptive slopes are proposed to improve the convergence rate of PINNs, but they cannot avoid the manually selection of activation function when solving different PDEs. 
We show that our method is orthogonal to these methods and can be extended by introducing
the adaptive slope into each function in the candidate set.  

Our work focuses on the influence of adaptive activation functions on PINNs optimization. The specific contributions of this paper can be summarized as: 
\begin{itemize}
    \item We shed light on the relationship between activation functions and the optimization difficulty of PDE-based constraint. Specifically, we highlight the sensitivity of PINNs to the choice of activation functions, which can be related to various characteristics of the PDE systems. We hope our work could inspire further study on the convergence issue of PINNs from the perspective of network architectures. 
    \item We explore the automatic design of activation functions for PINNs. We evaluate different adaptive activation functions on a series of PDEs, including Poisson's equation, convection equation, Burgers' equation, Allen-Cahn equation, Korteweg–de Vries equation, Cahn-Hilliard equation and Navier-Stokes equations. The limitations of existing methods are discussed in the context of PINNs.
    \item We tailor the simple idea of learning combinations of activation functions to the optimization of PINNs. The resulting method can be used to solve different types of PDE systems in an interpretable way. Extensive experiments demonstrate the effectiveness of the proposed method. We also present a comprehensive ablation study to provide a better understanding. We further attempt to explain the effectiveness of learning combinations of activation functions from the perspective of neural tangent kernel. 
\end{itemize}

The paper is organized as follows. In \cref{method}, we first give a brief introduction to physics-informed neural networks. Then, we investigate the influence of activation functions on PINNs on a simple benchmark problem with analytical solutions in \cref{motivation}. 
The results suggest that the choice of activation functions is crucial for PINNs and depends on the problem. Motivated by this observation, we propose to learn specialized activation functions for different PDE systems with adaptive activation functions as described in \cref{ada_af}. Subsequently, we compare different types of adaptive activation functions through extensive experiments in \cref{experiments}. Finally, we summarize our work and discuss the limitations in \cref{conclusion}. 

\section{Method}
\label{method}

\subsection{Physics-informed neural networks} 
Physics-informed neural networks have emerged as a promising method for solving forward and inverse problems of integer-order PDEs~\cite{raissi2019physics,chen2020physics,lu2021physics,karniadakis2021physics}, fractional PDEs~\cite{pang2019fpinns}, and stochastic PDEs~\cite{zhang2019quantifying}.
In this work, we consider PDEs of the general form 
\begin{equation}
    \vu_t + \mathcal{N}[\vu(t,\vx);\lambda]=0,\ \vx\in \Omega \subset \mathbb{R}^d,\ t\in[0,T],
\end{equation}
subject to the initial and boundary conditions 
\begin{align}
    \vu(0, \vx) &= \vu^0(\vx),\ \vx \in \Omega, \\
    \mathcal{B}[\vu]&=0,\ \vx \in \partial\Omega,\ t \in [0,T],
\end{align}
where $\vu(t,\vx)$ denotes the solution, $\mathcal{N}[\cdot; \lambda]$ is a differential operator parameterized by $\lambda$, $\mathcal{B}[\cdot]$ is a boundary operator, and subscripts denote the partial differentiation. 

A physics-informed neural network (PINN) $\vu'(t,\vx; \boldsymbol{\theta})$ is optimized to approximate the solution $\vu(t,\vx)$ by minimizing the following objective function
\begin{equation}
    \mathcal{L}(\boldsymbol{\theta}) = \mathcal{L_\mathrm{ic}}(\boldsymbol{\theta}) + \mathcal{L_\mathrm{bc}}(\boldsymbol{\theta}) + \mathcal{L_\mathrm{r}}(\boldsymbol{\theta}),
\end{equation}
where 
\begin{align}
    \mathcal{L_\mathrm{ic}}(\boldsymbol{\theta}) &= \frac{1}{N_\mathrm{ic}}\sum_{i=1}^{N_\mathrm{ic}}||\vu'(0,\vx_\mathrm{ic}^i;\boldsymbol{\theta})-\vu^0(\vx_\mathrm{ic}^i)||_2^2, \\
    \mathcal{L_\mathrm{bc}}(\boldsymbol{\theta}) &= \frac{1}{N_\mathrm{bc}}\sum_{i=1}^{N_\mathrm{bc}}||\mathcal{B}[\vu'(t_\mathrm{bc}^i,\vx_\mathrm{bc}^i;\boldsymbol{\theta})]||_2^2,\\
    \mathcal{L_\mathrm{r}}(\boldsymbol{\theta}) &= \frac{1}{N_\mathrm{r}}\sum_{i=1}^{N_\mathrm{r}}||\vu_t'(t_\mathrm{r}^i, \vx_\mathrm{r}^i; \boldsymbol{\theta})+\mathcal{N}[\vu'(t_\mathrm{r}^i, \vx_\mathrm{r}^i;\boldsymbol{\theta});\lambda]||_2^2 \label{eq:L_r}.
\end{align}
Here $\boldsymbol{\theta}$ denotes the weights and biases of neural networks. The $\mathcal{L_\mathrm{ic}}$ and $\mathcal{L_\mathrm{bc}}$ measure the prediction error on initial training data $\{\vx_\mathrm{ic}^i\}_i^{N_\mathrm{ic}}$ and boundary training data 
$\{t_\mathrm{bc}^i, \vx_\mathrm{bc}^i\}_i^{N_\mathrm{bc}}$. The residual loss $\mathcal{L_\mathrm{r}}$ is imposed to make the neural network satisfy the PDE constraint on a set of collocation points  $\{t_\mathrm{r}^i,\vx_\mathrm{r}^i\}_i^{N_\mathrm{r}}$. To compute the residual loss $\mathcal{L_\mathrm{r}}$, partial derivatives of the neural network output with respect to $t$ and $\vx$ can be obtained via automatic differentiation techniques~\cite{baydin2018automatic}. 

\subsection{Activation functions in PINNs}
\label{motivation}
Across the deep learning community, the ReLU enjoys widespread adoption and becomes the default activation function. However, there is no such default choice of activation functions for PINNs, regardless of the importance of activation functions in the optimization and expressivity of neural networks.
One of the reasons lies in the failure of ReLUs in PINNs, whose second-order derivative is zero everywhere. 
More importantly, PINNs show great sensitivity to activation functions due to the various characteristics of the underlying PDE system. To give a further illustration, we compare the performance of PINNs with different activation functions on the following toy examples. 

\paragraph{Problem formulation} Here we consider the 1D Poisson's equation as 
\begin{equation}
    \Delta u(x) = f(x),\ x \in [0, \pi],
\end{equation}
with boundary condition $u(0)=0$ and $u(\pi)=\pi$, where $\Delta$ is the Laplace operator and $f(x)$ denotes the source term. As shown in \cref{solution_poisson}, we use different source terms to construct analytical solutions with various characteristics. For the problem \textbf{P1}, the solution is composed of trigonometric functions with different periods; for the problem \textbf{P2}, its solution is dominant by the exponential function. Following the setting described in~\cite{yu2022gradient}, we impose the boundary conditions as hard-constraints by choosing the surrogate of solution as 
\begin{equation}
    \hat u(x)= x(x-\pi)u'(x;\theta)+x,
\end{equation}
where $u'(x;\theta)$ is the output of PINN. Hence, the training objective is reduced to the residual loss $\mathcal{L_\mathrm{r}}$.  Note that we consider an extreme case of insufficient collocation points to highlight the difference between activation functions. 

For each problem, we use a 4-layer fully-connected networks with the hidden dimension set to 16. The model is initialized by the Xavier initialization~\cite{glorot2010understanding} and trained for 20000 iterations using the Adam optimizer~\cite{kingma2014adam}. We initialize the learning rate as 1e-3 and adapt it with a half-cycle cosine decay schedule. We use 32 collocation points for both \textbf{P1} and \textbf{P2}. 
For each problem, we experiment with 8 standard activation functions as described in \cref{sec:standard_act}, including trigonometric functions ($\mathrm{sin}$ and $\mathrm{cos}$), the hyperbolic tangent function ($\mathrm{tanh}$), the logistic function ($\mathrm{sigmoid}$), the Softplus function ($\mathrm{Softplus}$)~\cite{dugas2000incorporating,glorot2011deep}, the Exponential Linear Unit ($\mathrm{ELU}$)~\cite{clevert2015fast}, the Gaussian Error Linear Unit ($\mathrm{GELU}$)~\cite{hendrycks2016bridging} and the Swish function ($\mathrm{Swish}$)~\cite{hendrycks2016bridging,ramachandran2017searching,elfwing2018sigmoid}.
We repeat each experiment 10 times and report the mean and standard deviation of the $L_2$ relative error.

\begin{table}[htbp]
  \caption{The source terms and analytical solutions for the 1D Poisson's equation.}
  \label{solution_poisson}
  \centering
  \begin{tabular}{l|ll}
    \toprule
    Problems    & $f(x)$     & $u(x)$ \\
    \midrule
    \textbf{P1}  & $-(\sum_{i=1}^{20} i\mathrm{sin}(ix))$ & $x+\sum_{i=1}^{20} \mathrm{sin}(ix)/i$ \\
    \textbf{P2}  & $(x^3-2\pi x^2+(\pi^2+3)x-2\pi)e^{\frac{(x-\pi)^2}{2}}$     & $xe^{\frac{(x-\pi)^2}{2}}$  \\
    \bottomrule
  \end{tabular}
\end{table}

\begin{figure*}[ht]
\subfigure[\textbf{P1}]{
\begin{minipage}{.48\textwidth}
    \centering
    \includegraphics[width=1.0\textwidth]{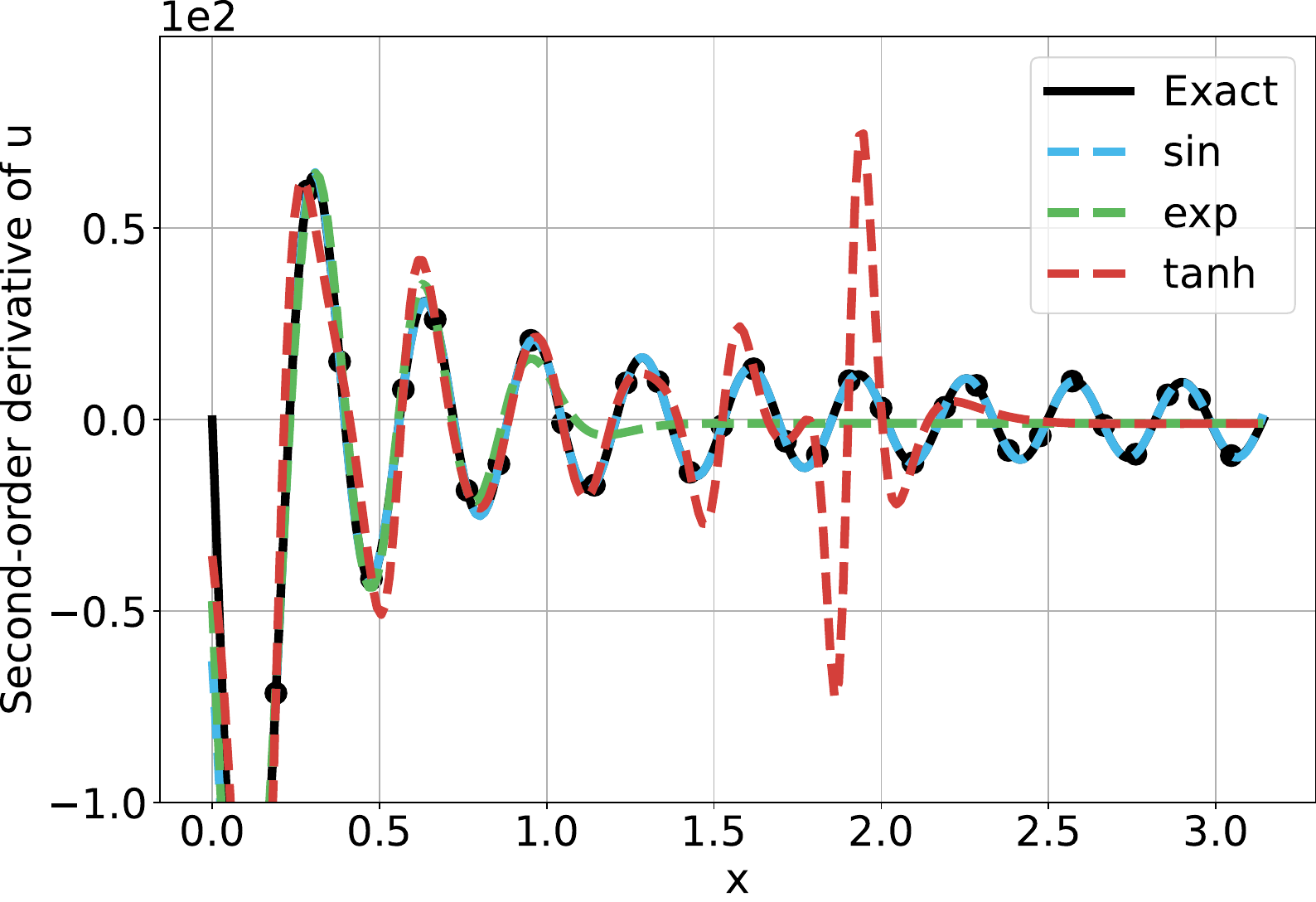}
\end{minipage}}
\subfigure[\textbf{P2}]{
\begin{minipage}{.48\textwidth}
    \centering
    \includegraphics[width=1.0\textwidth]{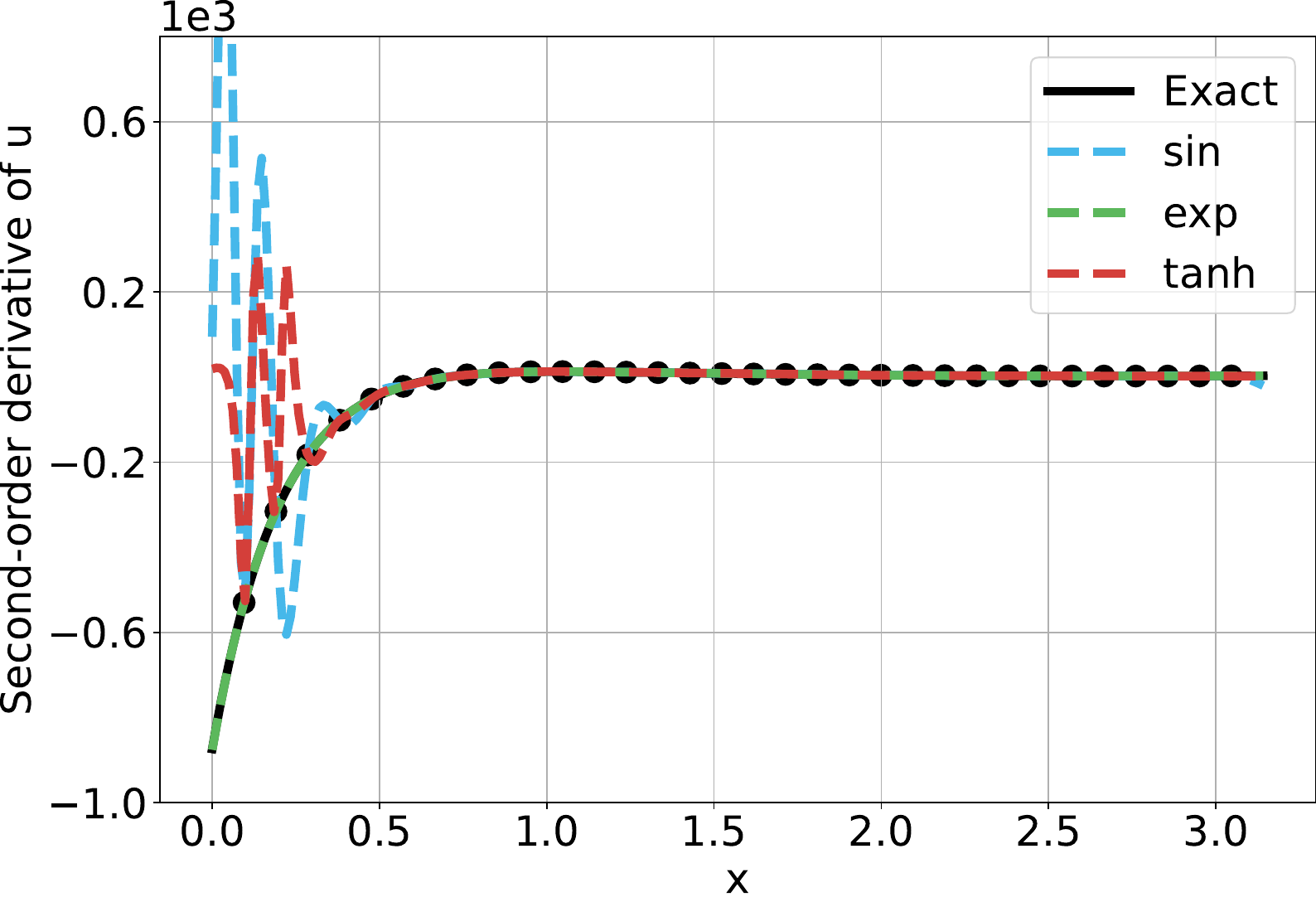}
\end{minipage}}
    \caption{Visualization of the residual loss $\mathcal{L_\mathrm{r}}$. In the case of 1D Poisson's equation, the residual loss $\mathcal{L_\mathrm{r}}$ penalizes the deviation of the second-order derivative of PINNs to the exact values on sampled collocation points (black dots in figures). We plot the results of $\mathrm{sin}$, $\mathrm{exp}$ and $\mathrm{tanh}$.}
    \label{fig:poisson_second_order_derivative}
    \vskip -0.15in
\end{figure*}

\textbf{The choice of activation functions affects the optimization of PINNs.} \cref{tab:error_poisson} shows the results of different activation functions on \textbf{P1} and \textbf{P2}. As we can see, the choice of activation functions makes a big difference on the prediction accuracy. For example, there exists a large performance gap between the $\mathrm{exp}$ and $\mathrm{tanh}$ functions on \textbf{P2} ($0.73\pm2.08\%$ vs. $75.77\pm7.93\%$). 
In \cref{fig:poisson_second_order_derivative}, we visualize the second-order derivative of PINNs to analyse the influence of activation functions on the optimization of residual loss $\mathcal{L_\mathrm{r}}$. We find that the commonly-used $\mathrm{tanh}$ function suffers from severe overfitting when solving these two problems with insufficient collocation points. In contrast, the optimization difficulty can be significantly reduced under the same conditions by using appropriate functions, such as the $\mathrm{sin}$ function for \textbf{P1} and the $\mathrm{exp}$ function for \textbf{P2}. 

\textbf{The optimal activation function depends on the property of PDE system.} As we can see in \cref{tab:error_poisson}, the $\mathrm{sin}$ and $\mathrm{exp}$ functions achieve the best performance to model the periodic nature of \textbf{P1} and the rapid decay of \textbf{P2}, respectively. This is because they can embed similar properties into the neural networks. Note that neither of them could achieve the best performance simultaneously on these two problems, due to the difference in the characteristics of the problems. One can find that while $\mathrm{sin}$ achieves the lowest error on \textbf{P1}, it produces poor results on \textbf{P2}. A similar phenomenon is observed on the $\mathrm{exp}$ function. This finding demonstrates the necessity of careful selection of activation functions when solving different problems. 

\begin{table}[htbp]
  \caption{Comparisons between different activation functions on 1D Poisson's equation. $L_2$ relative error (\%) is reported.}
  \label{tab:error_poisson}
  \centering
  \begin{tabular}{c|cccc}
    \toprule
    Problems    & $\mathrm{sin}$ & $\mathrm{exp}$ & $\mathrm{tanh}$ & $\mathrm{sigmoid}$\\
    \midrule
    \textbf{P1}  & $\bm{0.91\pm0.30}$ & $39.11\pm31.32$ & $34.27\pm45.21$ & $23.51\pm17.82$  \\
    \textbf{P2}  & $49.42\pm4.75$ & $\bm{0.73\pm2.08}$ & $75.77\pm7.93$ & $56.36\pm10.81$ \\
    \midrule
    Problems    & $\mathrm{Softplus}$ & $\mathrm{ELU}$ & $\mathrm{GELU}$ & $\mathrm{Swish}$\\
    \midrule
    \textbf{P1}  & $86.03\pm64.79$ & $36.68\pm20.61$ & $19.48\pm18.50$ & $33.16\pm36.81$  \\
    \textbf{P2}  & $\bm{0.65\pm0.23}$ & $40.96\pm97.75$ & $2.70\pm3.41$ & $1.05\pm0.42$ \\
    \bottomrule
  \end{tabular}
\end{table}

\subsection{Learning specialized activation functions for solving PDEs}
\label{ada_af}

In light of the diversity and complexity of PDEs, it is critical to select proper activation functions in PINNs. Existing works usually make the choice though trial-and-error. 

However, this strategy is inefficient especially when solving complex problems, where searching for a set of activation functions is necessary to make accurate predictions.
For instance, a combination of the sinusoidal and the exponential function is demonstrated effective to solve the heat transfer equation,
whose solution is periodic in space and exponentially decaying in time~\cite{zobeiry2021physics}. In this case, the trial-and-error strategy leads to the combinatorial search problem, which becomes infeasible when the candidate activation function set is large.
To address this problem, we introduce adaptive activation functions into PINNs to search for the optimal functions for different problems automatically. 

As we mentioned before, the major difference of existing adaptive activation functions lies in their search space. Here we consider four types of search space which could provide higher-order derivatives. Their formulations are as follows. 
\begin{itemize}
    \item \textbf{Linear combinations of common activation functions}~\cite{dushkoff2016adaptive,qian2018adaptive,manessi2018learning,sutfeld2020adaptive}: instead of categorically selecting one particular activation function, this kind of methods constructs the search space to be a linear combination of all candidate activation functions with learnable coefficients. We use the formulation from the Adaptive Blending Unit (ABU)~\cite{sutfeld2020adaptive}, because it covers different normalization methods for the coefficients. This formulation is presented as:
    \begin{equation}
        f(x) = \sum_{i=1}^N G(\alpha_i)\sigma_i(x),
    \end{equation}
    where $\sigma_i(\cdot)$ and $\alpha_i$ denote a candidate activation function and a learnable parameter, respectively. Together with a gate function $G(\cdot)$, the parameter $\alpha_i$ determines the weight (or coefficient) of its corresponding candidate activation function. Different gate functions correspond to different normalization methods in ABU. By default, we use the softmax function $G(\alpha_i) = \mathrm{exp}({\alpha_i})/\sum_{j=1}^N \mathrm{exp}({\alpha_j})$. In this case, the search space is restricted to the convex hull of the set of candidate activation functions.  
    
    \item \textbf{Learnable functions based on Taylor approximation}: the SLAF~\cite{goyal2019learning} proposes to approximate the optimal activation functions for different tasks with Taylor polynomial bases, which is presented as:
    \begin{equation}
        f(x)=\sum_{i=0}^ma_ix^i,
    \end{equation}
    where m is the order of Taylor series and is set to 5 in our experiments. The search space of SLAF covers a wind range of existing activation functions which could be accurately approximated by Taylor polynomial bases. The learnable parameters $\{a_i\}_{i=0}^m$ are initialized from a normal distribution.
    
    \item \textbf{Learnable functions based on Padé approximation}: Padé Activation Units (PAU) adopt the Padé approximation to construct its search space~\cite{molina2019pad}, which is presented as rational functions of the form
    \begin{equation}
        f(x)=\frac{\sum_{i=0}^ma_ix^i}{1+|\sum_{j=1}^nb_jx^j|},
    \end{equation}
    where m is set to 5 and n is set to 4 following the default setting in \cite{molina2019pad}. As an alternative to Taylor approximation, the Padé approximant often provides a better approximation of a given activation function. Note that the absolute value in its formulation is to avoid a zero-valued denominator. The learnable parameters $\{a_i\}_{i=0}^m$ and $\{b_j\}_{j=1}^n$ are initialized from a normal distribution.
    \item \textbf{Approximators for the general Maxout family}: motivated by the fact that the Swish function is a smooth approximation to ReLU, the ACON~\cite{ma2021activate} is proposed as an approximator to the general Maxout family activation functions. Its formulation is presented as 
    \begin{equation}
        f(x)=(p_1-p_2)x*\mathrm{sigmoid}(\beta(p_1-p_2)x)+p_2x,
    \end{equation}
    where $p_1$, $p_2$ and $\beta$ are learnable parameters. It is a smooth approximator to the $max(p_1x,p_2x)$ function. By fixing $p_1=1$ and $p_2=0$, the ACON is reduced to the Swish function. Compared with other adaptive activation functions, the search space of ACON has a limited degree of flexibility. The learnable parameters $p_1$, $p_2$ and $\beta$ are initialized from a normal distribution.
\end{itemize}

Since previous works mainly focus on the convolutional neural networks for the image classification task, learning adaptive activation functions in the context of PINNs faces new challenges in terms of smoothness and diversity, which arises from the peculiar nature of the problem. For example, the polynomial-based SLAF is subject to the exploding values and oscillation problems, which may hinder the optimization of PDE-based constraints. We also find that the PAU suffers from the training instability, because its higher-order derivatives are not continuous. While ACON meets the requirement of smoothness, its search space has limited diversity, which limits its performance in solving different PDEs with various characteristics.

To remove obstacles to apply adaptive activation functions to PINNs, we propose to tailor the idea of learning combinations of activation functions to the optimization of PDE-based constraints.
Taking the ABU~\cite{sutfeld2020adaptive} as an example, its search space is built upon the candidate functions set $\mathcal{F}=\{\sigma_1, \sigma_2,...,\sigma_N\}$. First, we remove those activation functions which cannot provide continuous and non-zero higher-order derivatives, such as ReLU, ELU and identity function. Second, we incorporate elementary functions with different properties according to our prior knowledge of a physics system at hand. For example, the sinusoidal and exponential functions can be leveraged to model the periodicity and rapid decay phenomena, which are difficult to model by neural networks with normal activation functions such as $\mathrm{tanh}$ as shown in \cref{motivation}.  
Note that none of the existing works~\cite{dushkoff2016adaptive,qian2018adaptive,manessi2018learning,sutfeld2020adaptive} have considered adding these functions to their candidate set, because there is no such prior knowledge to leverage in general domains like image classification or natural language processing. In contrast, we reveal that adaptive activation functions can be more powerful and interpretable in handling PDE systems. In addition to these elementary functions, we also include most of commonly-used activation functions to ensure the diversity. Finally, we further enhance the search space with the adaptive slope, which is proposed to improve the convergence rate of standard activation functions for PINNs~\cite{jagtap2020adaptive,jagtap2020locally}. This can be achieved by simply introducing a learnable scaling factor $\beta_i$ for each candidate function $\sigma_i$, which can be represented as 
\begin{equation}
    f(x) = \sum_{i=1}^N G(\alpha_i)\sigma_i(\beta_ix).
\end{equation}
We denote the modified version of ABU as ABU-PINN to highlight the special treatment tailored for PINNs. 

With adaptive activation functions, we formulate the manual selection of activation functions as a learning problem. We employ these methods by replacing all standard activation functions in PINNs. They can be applied in a neuron-wise manner where learnable parameters are allowed to be vary across neurons, or in a layer-wise manner where the parameters are shared for all neurons in one layer. Their learnable parameters can be optimized with other weights and biases in the networks through backpropagation to minimize the loss functions. Therefore, the neural networks learn to adapt the shape of activation functions to the PDE system during the training process. Besides the efficient optimization, the continuous parameterization of search space enables the discovery of novel functions which are different from the hand-designed ones.

\section{Experiments}
\label{experiments}
In this section, we first revisit the 1D Poisson's equation to show ABU-PINN can solve different types of problems in an interpretable way in \cref{exp_poisson}. Next, we compare the performance of different adaptive activation functions on several benchmarks of 1D time-dependent PDEs in \cref{exp_1d_time_dependent}, as well as 2D PDEs in \cref{exp_2d_pde}. Then, \cref{ablation} provides detailed ablation results to analyse the proposed {\name}. Finally, in \cref{exp_ntk}, we present an intuitive interpretation of the performance gain of ABU-PINN from the neural tangent kernel perspective.  

\subsection{Experiments on 1D Poisson's equation}
\label{exp_poisson}

First, we use the 1D Poisson's equation with different source terms as described in \cref{motivation} to demonstrate that the proposed ABU-PINN could learn optimal activation functions for problems with different properties by incorporating the prior knowledge of the PDE systems. 
To advance the modeling of the
periodic nature of \textbf{P1} and the rapid decay of \textbf{P2} simultaneously, we add the $\mathrm{sin}$ and  $\mathrm{exp}$ into the candidate functions to reduce the training difficulty of PINNs. 
We define $\mathcal{F}=\{\mathrm{sin}, \mathrm{exp}\}$ for simplicity and expect ABU-PINN could adapt its attention over these two candidate functions to the problem.
The trainable parameters $\{\alpha_{\mathrm{sin}}, \alpha_{\mathrm{exp}}\}$ are initialized as zeros, which implies equal attention over all candidate functions. We follow the experiments setting described in \cref{motivation}, except that we replace all the fixed-shape activation functions as ABU-PINN. 

The results of {\name} and its two candidate functions are presented in \cref{tab:error_poisson_ours}. It can be observed that {\name} can achieve comparable accuracy with the $\mathrm{sin}$ on \textbf{P1} and outperform the $\mathrm{exp}$ on \textbf{P2}, despite the large performance gap between its two candidate functions. The results verify the effectiveness of ABU-PINN for solving PDE systems with different source terms by incorporating suitable candidate functions. We also report the values of learned coefficients in \cref{tab:coeff_poisson_ours}. For \textbf{P1}, {\name} pays more attention on the $\mathrm{sin}$ function as expected. For \textbf{P2}, we observe that the learned activation functions are different across layers. The $\mathrm{exp}$ is preferred only on the third layer, which still leads a performance gain compared with standard activation functions. The results indicate that ABU-PINN can learn useful combination coefficients for different problems. 

\begin{table}[htbp]
\centering
\caption{Comparisons of standard activation and {\name} on the 1D Poisson's equation. $L_2$ relative error (\%) is reported. We also report the learned coefficients $\alpha_\mathrm{sin}$ and $\alpha_\mathrm{exp}$.} 
\subtable[The $L_2$ relative error (\%)]
{
\centering
\label{tab:error_poisson_ours}
\begin{tabular}{c|ccc} 
\toprule
Problems          & $\mathrm{sin}$ & $\mathrm{exp}$ & {\name}  \\
\midrule
\textbf{P1} & $0.91\pm0.30$ & $39.11\pm32.32$ & $\bm{0.88\pm0.42}$\\
\textbf{P2} & $49.42\pm4.75$&  $0.73\pm2.08$ & $\bm{0.01\pm0.006}$ \\
\bottomrule
\end{tabular}
}
\subtable[The learned coefficients]
{
\centering
\label{tab:coeff_poisson_ours}
\resizebox{1.0\linewidth}{!}
  {
\begin{tabular}{c|cc|cc|cc} 
\toprule
 & \multicolumn{2}{c}{Layer-1} & \multicolumn{2}{c}{Layer-2} & \multicolumn{2}{c}{Layer-3}  \\
Problems & $\alpha_\mathrm{sin}$ & $\alpha_\mathrm{exp}$ & $\alpha_\mathrm{sin}$ & $\alpha_\mathrm{exp}$ & $\alpha_\mathrm{sin}$ & $\alpha_\mathrm{exp}$ \\
\midrule
\textbf{P1} & $0.81\pm0.02$ & $0.19\pm0.02$ & $0.75\pm0.06$ & $0.25\pm0.06$ & $0.84\pm0.04$ & $0.16\pm0.04$ \\
\textbf{P2} & $0.58\pm0.08$ & $0.42\pm0.08$ & $0.53\pm0.05$ & $0.47\pm0.05$ & $0.45\pm0.06$ & $0.55\pm0.06$ \\

\bottomrule
\end{tabular}
}
}
\end{table}

\subsection{Experiments on 1D time-dependent PDEs}
\label{exp_1d_time_dependent}
We provide a comprehensive comparison between different adaptive activation functions to solve various time-dependent PDEs, ranging from first-order linear PDE to fourth-order nonlinear PDE, including the convection equation, the Burgers' equation, the Allen-Cahn equation, the Korteweg–de Vries equation, and the Cahn-Hilliard equation. 

\subsubsection{Problem formulations}
\label{sec:problem_form}
\paragraph{Convection equation.}
Consider a one-dimensional linear convection problem as 
\begin{align}
    u_t+\beta u_x&=0,\ x\in[0,2\pi],\ t\in[0,1], \\ 
    u(0,x)&=\mathrm{sin}(x),\\
    u(t,0)&=u(t,2\pi). 
\end{align}
The solution of this problem is periodic over time, whose period is inversely proportional to the convection coefficient $\beta$. The previous work~\cite{krishnapriyan2021characterizing} finds it difficult for vanilla PINNs to learn the solution with a large $\beta$ and proposes to train the PINNs with curriculum learning. 
We show that accurate prediction can be achieved with a normal training strategy by using a suitable activation function. Moreover, the performance can be further improved by the proposed {\name}. Here we set the convection coefficient $\beta$ to 64. We use $N_{\mathrm{ic}}=512$ points for the initial condition and $N_{\mathrm{bc}}=200$ for the boundary condition. The residual loss $\mathcal{L_\mathrm{r}}$ is computed on $N_\mathrm{r}=6400$ randomly sampled collocation points.

\paragraph{Burgers' equation.}
We consider the one-dimensional Burgers' equation:
\begin{align}
    u_t + uu_x &= \nu u_{xx},\ x \in [-1,1],\ t \in [0,1],\\
    u(0,x) &= -\mathrm{sin}(\pi x),\\
    u(t,-1)&=u(t,1)=0,
\end{align}
where $\nu=0.01/\pi$. Following \cite{lu2021deepxde}, we uniformly sample $N_{\mathrm{ic}}=256$ and $N_{\mathrm{bc}}=100$ points for initial and boundary training data, respectively. We compute the residual loss $\mathcal{L_\mathrm{r}}$ on $N_\mathrm{r}=4800$ collocation points, which are randomly selected from the spatial-temporal domain.

\paragraph{Allen-Cahn equation.}
We next consider the Allen-Cahn equation as 
\begin{align}
    u_t = D u_{xx} + 5(u&-u^3),\ x \in [-1,1],\ t \in [0,1],\\
    u(0,x) &= x^2\mathrm{cos}(\pi x),\\
    u(t,-1)&=u(t,1)=-1,
\end{align}
where $D=0.001$ following the setting in \cite{yu2022gradient}. We choose the following surrogate of solution to enforce the initial and boundary conditions: 
\begin{equation}
    \hat u(t,x)= x^2\mathrm{cos}(\pi x)+t(1-x^2)u'(t,x;\theta),
\end{equation}
where $u'(t,x;\theta)$ is the output of neural networks. We use $N_\mathrm{r}=8000$ collocation points. 

\paragraph{Korteweg–de Vries equation.} We consider the Korteweg–de Vries (KdV) equation as
\begin{align}
    u_t+\lambda_1 uu_x + \lambda_2 u_{xxx}&=0,\ x\in[-1,1],\ t\in[0,1], \\ 
    u(0,x)&=\mathrm{cos}(\pi x),\\
    u(t,-1)=u(t,1)&,\ u_x(t,-1)=u_x(t,1),
\end{align}
where $\lambda_1=1$ and $\lambda_2=0.0025$ following the setting in \cite{raissi2019physics}. We choose the following surrogate of solution to enforce the initial conditions: 
\begin{equation}
    \hat u(t,x)= \mathrm{cos}(\pi x)+tu'(t,x;\theta),
\end{equation}
where $u'(t,x;\theta)$ is the output of neural networks. We use $N_{\mathrm{bc}}=200$ training points for the boundary condition and $N_\mathrm{r}=8000$ collocation points for the residual loss $\mathcal{L_\mathrm{r}}$.

\paragraph{Cahn-Hilliard equation.} We consider the phase space representation of Cahn-Hilliard equation as
\begin{align}
    u_t-\nabla^2(-\lambda_1 \lambda_2 h+\lambda_2(u^3-u)) &=0,\ h=\nabla^2u,\ x\in[-1,1],\ t\in[0,1], \\ 
    u(0,x)=\mathrm{cos}(\pi x)&-\mathrm{exp}(-4(\pi x)^2), \\
    u(t,-1)=u(t,1)&,\ u_x(t,-1)=u_x(t,1),\\ 
    h(t,-1)=h(t,1)&,\ h_x(t,-1)=h_x(t,1),
\end{align}
where $\lambda_1=0.02$ and $\lambda_2=1$ following the setting in \cite{mattey2022novel}. The previous work proposes a novel sequential training method to solve this strongly non-linear and high-order problem \cite{mattey2022novel}. We find that a normal training strategy can lead to a comparable result if we add more collocation points around $t=0$. To be specific, we sample 4000 collocation points from time interval $[0,0.05)$ and 8000 points from $[0.05,1]$. We set $N_{\mathrm{ic}}=256$ and $N_{\mathrm{bc}}=100$. We show that PINNs under this training setting can work as a strong baseline and can be further improved by {\name}.

\subsubsection{Main results}
\label{1d_main_results}
\paragraph{Experiment setups.} We train the PINNs by a two-stage optimization except the convection equation. At the first stage, the model is trained for a certain number of iterations by Adam~\cite{kingma2014adam}. Then, the L-BFGS~\cite{byrd1995limited} is used to train the network until convergence. The first stage is to provide a good initialization for the L-BFGS optimizer. In the case of convection equation, only the Adam optimizer is used. The detailed experimental configurations can be found in \cref{tab:config}. 
The model is initialized by the Xavier initialization. The learning rate is adapted with a half-cycle cosine decay schedule in the first stage with Adam optimizer. We repeat each experiment 5 times and report the mean and standard deviation of the $L_2$ relative error. We run all experiments on one NVIDIA GeForce RTX 2080Ti GPU.
\begin{table}[htbp]
  \caption{The experimental configurations for 1D time-dependent problems.}
  \label{tab:config}
  \centering
  \resizebox{1.0\linewidth}{!}
 {
  \begin{tabular}{c|cc|cc|c|ccc}
    \toprule
    & \multicolumn{2}{c}{Network} & \multicolumn{2}{c}{Adam} & \multicolumn{1}{c}{L-BFGS} & \multicolumn{3}{c}{Data}\\
    Problems    & Depth & Width & Iters & lr & Max Iters  & $N_\mathrm{ic}$ & $N_\mathrm{bc}$ & $N_\mathrm{r}$ \\
    \midrule
    Convection equation & 6 & 64 & 100k & 2e-3 & -  & 512 & 200 & 6400 \\
    Burgers' equation & 4 & 32 & 15k & 1e-3 & 15k  & 256 & 100 & 5300 \\
    Allen-Cahn equation & 4 & 32 & 40k & 1e-3 & 15k  & - & - & 8000 \\
    KdV equation & 4 & 32 & 40k & 1e-3 & 15k &  - & 200 & 8000 \\
    Cahn-Hilliard equation & 4 & 32 & 100k & 1e-3 & 15k  & 256 & 100 & 15000 \\
    \bottomrule
  \end{tabular}
  }
\end{table}

\begin{table}
\centering
\caption{Comparisons of standard activation functions, adaptive activation functions and their counterparts with adaptive slopes (AS) on 1D time-dependent PDEs. We report the $L_2$ relative error (\%) for each problem and the average error rate over all problems. The better results are \textbf{bold-faced}.}
\label{tab:error_t_pde}
\resizebox{1.0\linewidth}{!}
 {
\begin{tabular}{c|ccccc|c} 
\toprule
 & Convection  & Burgers' & Allen-Cahn &  KdV  & Cahn-Hilliard & Average\\
 Method & equation &  equation & equation & equation & equation & error \\
\midrule
$\mathrm{sin}$ & $0.36\pm0.15$ & $5.18\pm3.73$& $3.57\pm0.64$ &  $0.63\pm0.17$ & $1.72\pm0.61$ & 2.29\\
$\mathrm{tanh}$ & $6.83\pm4.79$ & $0.26\pm0.13$& $1.34\pm0.54$ &  $1.32\pm1.12$ & $4.02\pm4.56$ & 2.75\\
$\mathrm{sigmoid}$& $70.38\pm2.98$ & $1.24\pm1.05$ & $1.63\pm0.13$ &  $2.34\pm0.53$ & $3.12\pm2.72$ & 15.74\\
$\mathrm{GELU}$ & $39.29\pm35.51$ & $4.43\pm2.87$ & $3.93\pm0.78$ & $1.21\pm0.41$ & $1.01\pm1.27$ & 9.97 \\
$\mathrm{Swish}$ & $5.59\pm2.18$ & $8.27\pm4.35$ & $5.56\pm1.28$ & $1.73\pm0.10$ & $2.22\pm2.60$ & 4.67\\
$\mathrm{Softplus}$& $55.39\pm2.46$ & $17.75\pm8.11$ & $17.72\pm6.04$ & $6.23\pm0.44$ & $9.67\pm2.57$ & 21.35\\
$\mathrm{ELU}$& $6.67\pm0.96$ & $46.34\pm2.36$ & $52.55\pm2.95$ & $78.95\pm2.57$ & $90.77\pm2.07$ & 55.66\\
\midrule
SLAF & $0.36\pm0.18$ & $43.93\pm0.66$ & $33.93\pm9.97$ & $25.23\pm1.28$ & $52.57\pm27.93$ & 31.20\\
PAU & $45.78\pm35.47$ & $48.31\pm8.21$ & $43.83\pm15.18$ & $68.11\pm13.49$ & $115.59\pm3.79$ & 64.32\\
ACON & $3.55\pm1.66$ & $1.18\pm1.55$ & $3.88\pm1.82$ & $1.52\pm0.35$ & $2.46\pm1.96$ & 2.52\\
{\textbf{\name}} &$\bm{0.06\pm0.03}$ &$\bm{0.21\pm0.11}$ & $\bm{0.76\pm0.26}$ & $\bm{0.34\pm0.05}$ & $\bm{0.50\pm0.15}$ & $\bm{0.37}$ \\
\midrule 
With AS \\
\midrule
$\mathrm{sin}$& $0.28\pm0.08$ & $1.82\pm1.58$ & $3.57\pm0.46$ &  $0.57\pm0.19$ & $1.73\pm0.73$ & 1.59 \\
$\mathrm{tanh}$& $3.06\pm1.65$ & $0.16\pm0.09$ & $0.81\pm0.21$ & $1.71\pm1.53$ & $2.11\pm0.49$ & 1.57 \\
$\mathrm{GELU}$& $38.00\pm39.53$ & $0.71\pm0.38$ & $1.99\pm0.63$ & $0.79\pm0.17$ & $0.96\pm0.57$ & 8.49\\
{\textbf{\name}}& $\bm{0.05\pm0.02}$ & $\bm{0.15\pm0.10}$ & $\bm{0.58\pm0.15}$ & $\bm{0.34\pm0.08}$ & $\bm{0.35\pm0.07}$ & $\bm{0.29}$\\
\bottomrule
\end{tabular}
}
\end{table}

\paragraph{Setups of standard and adaptive activation functions.} We experiment with several commonly-used activation functions, including the sinusoidal functions ($\mathrm{sin}$), the hyperbolic tangent function ($\mathrm{tanh}$), the logistic function ($\mathrm{sigmoid}$), the Softplus function ($\mathrm{Softplus}$)~\cite{dugas2000incorporating,glorot2011deep}, the Exponential Linear Unit ($\mathrm{ELU}$)~\cite{clevert2015fast}, the Gaussian Error Linear Unit ($\mathrm{GELU}$)~\cite{hendrycks2016bridging} and the Swish function ($\mathrm{Swish}$)~\cite{ramachandran2017searching}. The details of each activation function can refer to \cref{sec:standard_act}. We compare the performance of adaptive activation functions which could provide higher-order derivatives, including SLAF, PAU, ACON and ABU-PINN. 
We employ all adaptive activation functions in a layer-wise manner by default. 
For SLAF, PAU and ACON, the learnable parameters are initialized from a normal distribution.
We set the candidate set $\mathcal{F}$ as $\{\mathrm{sin}, \mathrm{tanh}, \mathrm{GELU}, \mathrm{Swish}, \mathrm{Softplus}\}$ and initialize the learnable parameters $\{\alpha_i\}_{i=1}^{N}$ as zeros for {\name}. This candidate set includes the $\mathrm{sin}$ function to help the learning of periodicity and several commonly-used activation functions to ensure the diversity. We show that this candidate set achieves robust and consistent performances over these five time-dependent PDEs.
We also compare {\name} with standard activation functions with the layer-wise adaptive slopes~\cite{jagtap2020locally}. In this case, the scaling factors $\{\beta_i\}_{i=1}^{N}$ are initialized as ones.  

\begin{figure*}[htbp]
\subfigure{
\begin{minipage}{0.98\textwidth}
    \centering
    \includegraphics[width=1.00\textwidth]{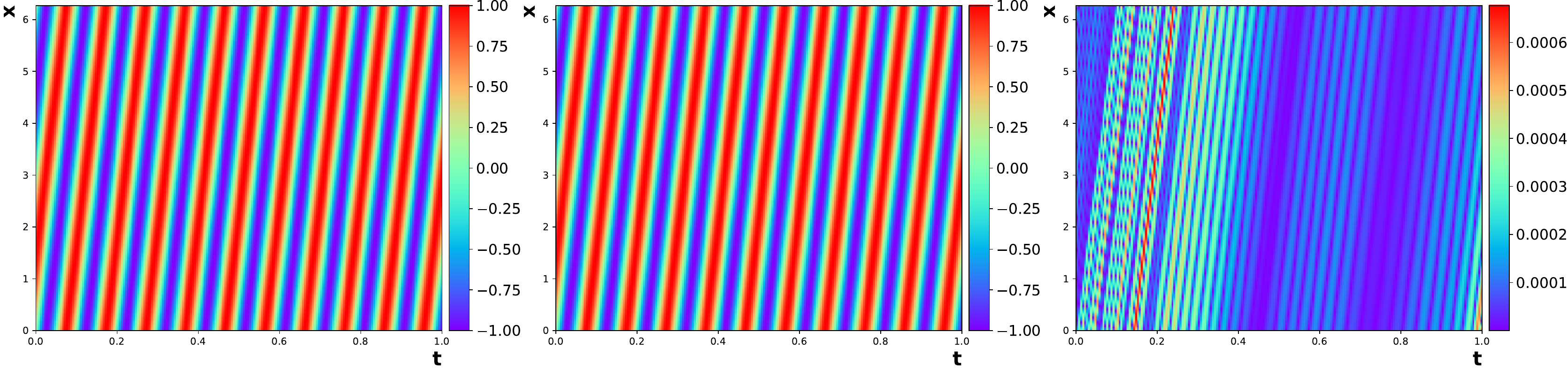}
\end{minipage}
}\\
\subfigure{
\begin{minipage}{0.98\textwidth}
    \centering
    \includegraphics[width=1.00\textwidth]{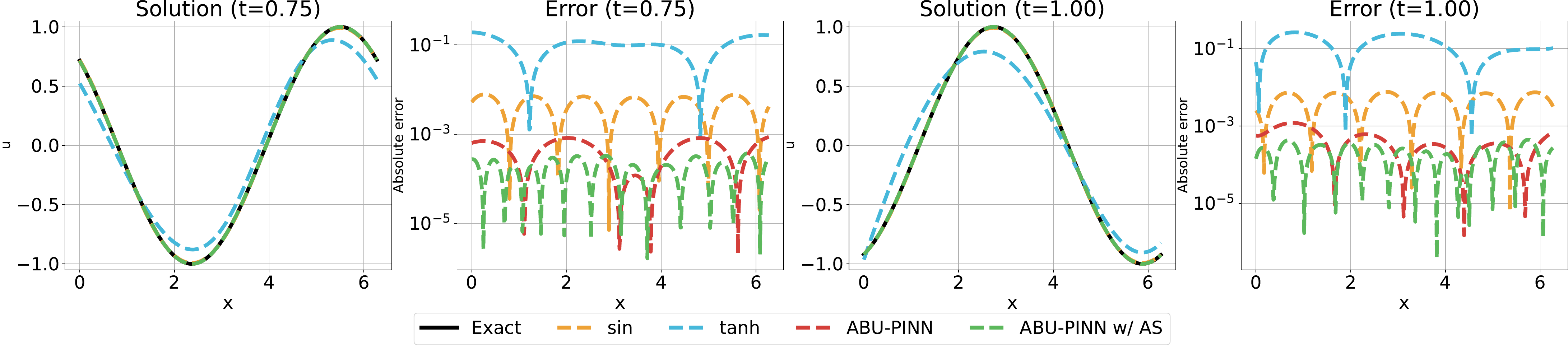}
\end{minipage}}
    \textcolor{black}{\caption{The convection equation. Top: the exact solution (left), the predictions of {\name} (middle), and the absolute error between them (right). Bottom: predicted solutions at different time snapshots.}\label{fig:pred_detailed_convection}}
    
\end{figure*}

\paragraph{Comparisons with fixed activation functions.} In \cref{tab:error_t_pde}, we presents the $L_2$ relative error of {\name} and seven standard activation functions on five time-dependent PDEs. One can observe that {\name} outperforms standard activation functions on all PDEs. For example, {\name} reduces the error rate by 83\% on the convetion equation ($0.06\pm0.03\%$ vs. $0.36\pm0.15\%$), by 46\% on the KdV equation ($0.34\pm0.05\%$ vs. $0.64\pm0.34\%$), and by 56\% on the Cahn-Hilliard equation ($0.50\pm0.15\%$ vs. $1.01\pm1.27\%$), compared with optimal standard activation function of each problem. More importantly, {\name} performs consistently over all these problems, while standard activation functions suffer from high variance in the performance of different PDEs.  
\textcolor{black}{We present the predicted solutions of {\name} for convection equation in \cref{fig:pred_detailed_convection} and Allen-Cahn equation in \cref{fig:pred_detailed_ac}. More visualization results can be found in Appendix \cref{fig:pred_detailed_burger,fig:pred_detailed_kdv,fig:pred_detailed_ch}.}

\begin{figure*}[htbp]
\subfigure{
\begin{minipage}{0.98\textwidth}
    \centering
    \includegraphics[width=1.00\textwidth]{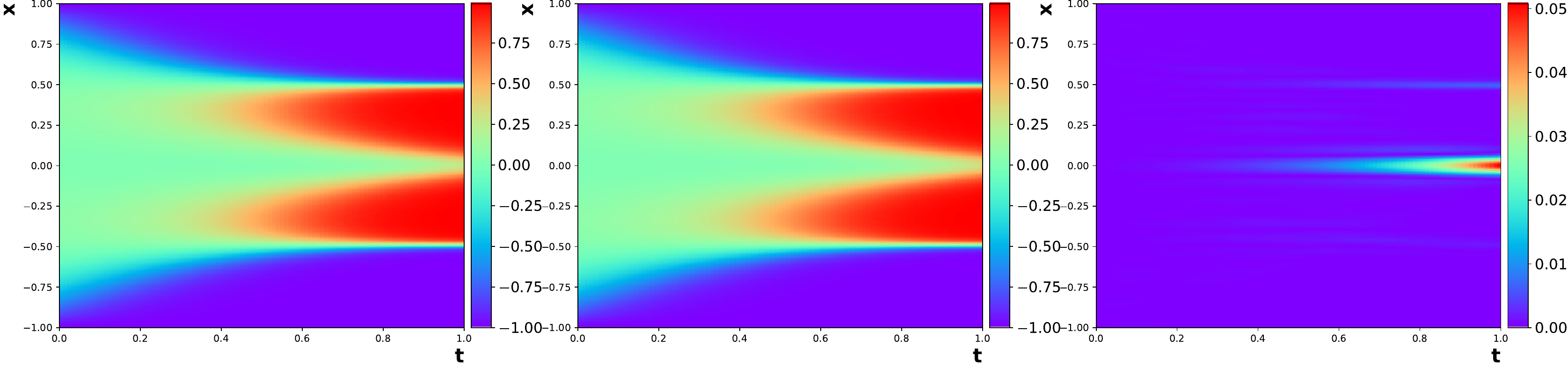}
\end{minipage}
}\\
\subfigure{
\begin{minipage}{0.98\textwidth}
    \centering
    \includegraphics[width=1.00\textwidth]{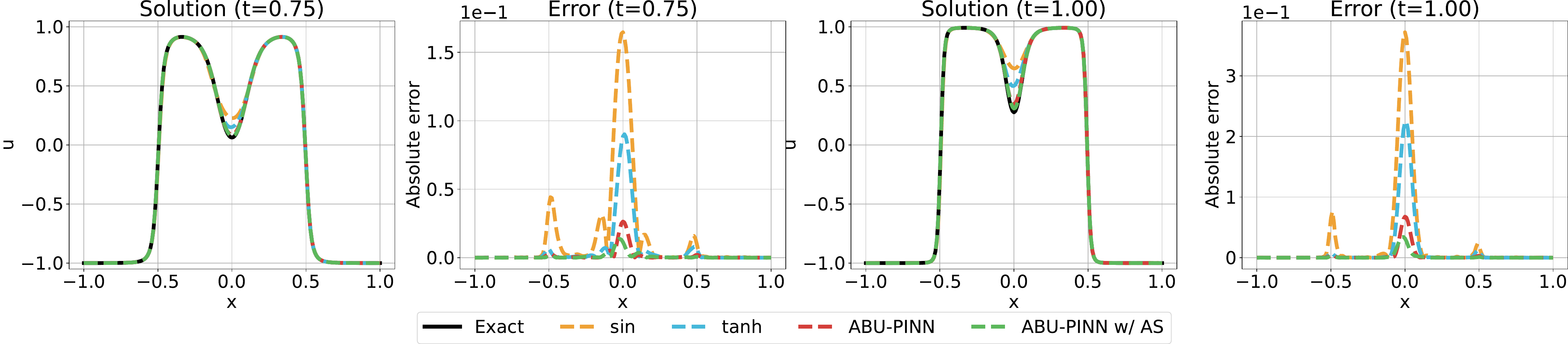}
\end{minipage}}
    \textcolor{black}{\caption{The Allen-Cahn equation. Top: the exact solution (left), the predictions of {\name} (middle), and the absolute error between them (right). Bottom: predicted solutions at different time snapshots.}\label{fig:pred_detailed_ac}}
    
\end{figure*}

\begin{figure*}[ht]
\subfigure[The Burgers' equation]{
\begin{minipage}{.49\textwidth}
    \centering
    \includegraphics[width=0.9\textwidth]{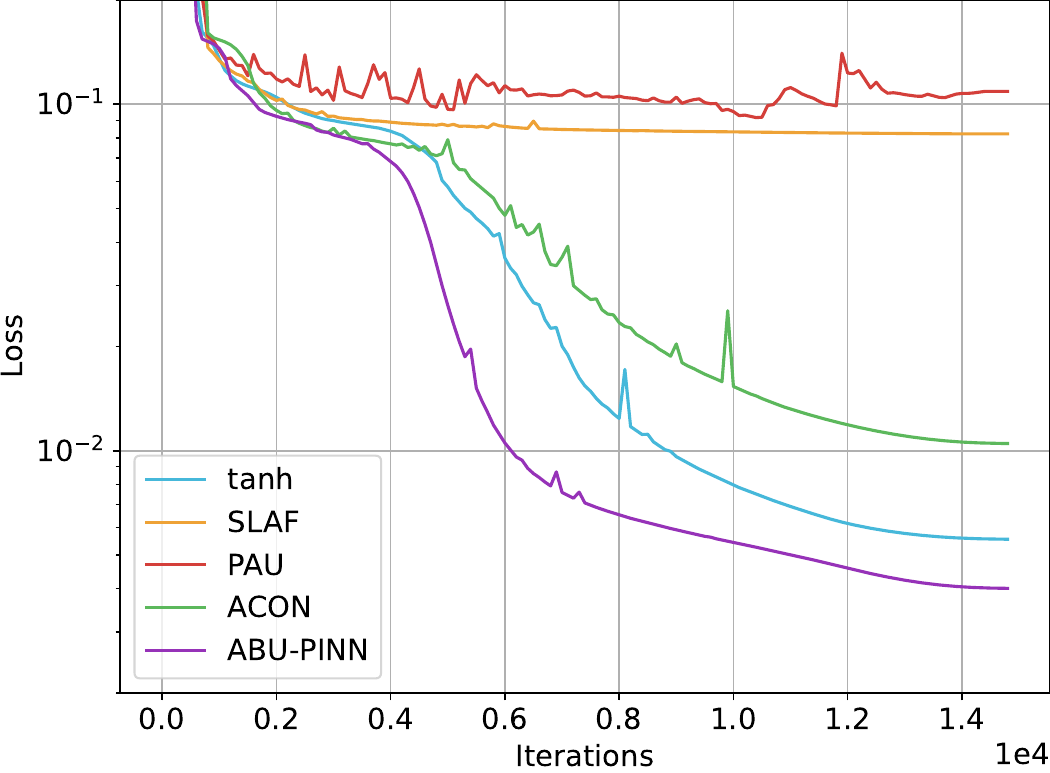}
\end{minipage}}
\subfigure[The Allen-Cahn equation]{
\begin{minipage}{.49\textwidth}
    \centering
    \includegraphics[width=0.9\textwidth]{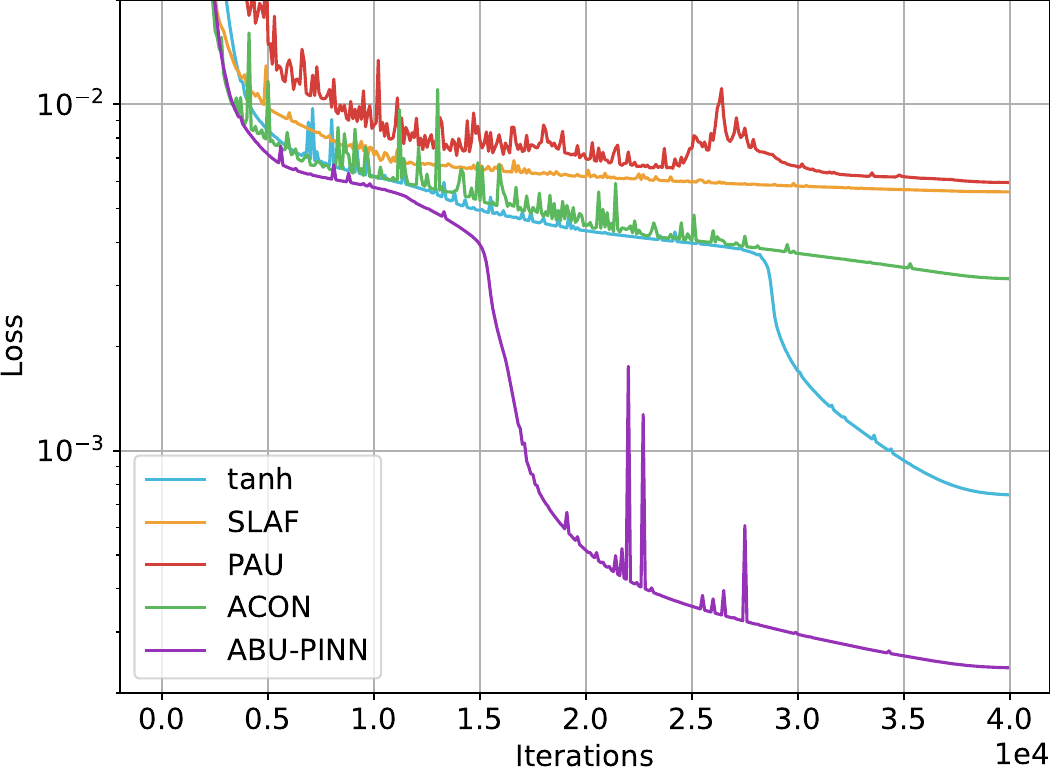}
\end{minipage}}
    \caption{The training losses of different activation functions against iterations.}
    \label{fig:training_loss}
    \vspace{-1.3em}
\end{figure*}

\paragraph{Comparisons with other adaptive activation functions.} The results of different adaptive activation functions are shown in \cref{tab:error_t_pde}. SLAF achieves performance comparable to the best standard activation function on the first-order convection equation ($0.36\pm0.18\%$ vs. $0.36\pm0.15\%$), but does not produce accurate predictions for other higher-order PDEs. Although SLAF shares a similar formulation with {\name}, its polynomial bases can cause vanishing or exploding gradients due to the high order powers in its derivatives. This problem might hinder the optimization of higher-order PDE-based constraints. One can observe that PAU performs poorly on all PDEs. To figure out the reason, we plot the training loss in \cref{fig:training_loss} and find PAU suffers from instability of training. We attribute the training instability of PAU to its discontinuous derivatives, which arise from the absolute value in the denominator of its formulation. ACON achieves stable performance but does not surpass the best standard activation function in each problem due to the limited flexibility of its formulation. 
The proposed {\name} surpasses these methods by leveraging its diverse and smooth candidate functions.

\paragraph{Comparisons with activation functions with adaptive slopes.} We also compare {\name} with three standard activation function enhanced with adaptive slopes (AS). We find {\name} without AS achieves competitive or even better results. For example, {\name} without AS outperforms the $\mathrm{tanh}$ with AS on the Allen-Cahn equation  ($0.76\pm0.26\%$ vs. $0.81\pm0.21\%$) and outperforms the $\mathrm{sin}$ with AS on the convection equation ($0.06\pm0.03\%$ vs. $0.28\pm0.08\%$). The performance of {\name} can be further improved by incorporating adaptive slopes into the search space. As a showcase, {\name} with AS obtains a performance gain by 64\% compared to the $\mathrm{GELU}$ with AS on the Cahn-Hilliard equation and by 82\% compared to the $\mathrm{sin}$ with AS on the convection equation. Furthermore, our method has a better generalization ability and achieves the lowest average error rate over all problems (0.29\% with AS and 0.37\% without AS), while the adaptive slopes cannot alleviate the need of manual selections of activation functions. 

\textcolor{black}{\paragraph{Computational cost.} We present the computational cost of each activation function in \cref{tab:comp_cost} and Appendix \cref{fig:training_loss_time_log}. The computational cost of {\name} is 3.1 times that of standard activation functions and 1.5 times that of ACON. Despite the incurred overhead, the proposed {\name} attains a significant improvement (approximately 85\%) in the averaged performance of five one-dimensional time-dependent PDEs. Moreover, {\name} surpasses the other two adaptive activation functions, namely SLAF and PAU, while simultaneously maintaining a comparable computational cost.}

\begin{table}
\color{black}
\centering
\caption{The computational cost of each activation function. We report the wall-clock time required to train each network for 10000 iterations in the case of Allen-Cahn equation, conducted on a NVIDIA GeForce RTX 2080Ti GPU. }

\label{tab:comp_cost}
\begin{tabular}{c|ccccc} 
\toprule
Method & $\mathrm{tanh}$ & SLAF & PAU & ACON & {\name} \\
\midrule
Time (sec) & 250 & 720 & 960 & 520 & 770 \\
\bottomrule
\end{tabular}
\color{black}
\end{table}

\paragraph{Visualization of the learned {\name}.} We show the learned {\name} in \cref{fig:learned_piac}. One can observe the learned activation functions are different from candidate functions. We also find differences in learned functions of different layers. For example, a deeper layer tends to has a larger weight of $\mathrm{sin}$ function in the case of convection equation. Moreover, the learned functions vary across problems, which conforms to our motivation to learn specialized activation functions for different PDE systems. 

\color{black}
\subsection{{Experiments on 2D PDEs}}
\label{exp_2d_pde}

\subsubsection{Problem formulations.} This section evaluates the efficacy of {\name} on two well-established benchmark problems in the field of computational fluid dynamics, which are governed by the two-dimensional incompressible Navier-Stokes equations.

\paragraph{Flow in a lid-driven cavity.} This case involves the simulation of fluid motion inside a square domain, where the top lid is driven with a constant velocity while the remaining boundaries are considered impermeable and stationary. This setup creates a confined cavity where the flow dynamics are determined by the steady-state Navier-Stokes equations as 
\begin{align}
    \lambda_1(uu_x+vu_y)+p_x-\lambda_2(u_{xx}+u_{yy})&=0,\ &(x,y)\in\Omega,\\
    \lambda_1(uv_x+vv_y)+p_y-\lambda_2(v_{xx}+v_{yy})&=0,\ &(x,y)\in\Omega,\\ 
    u_x+v_y&=0, \ &(x,y)\in\Omega,\\
    u(x,y) = 1,\ v(x,y) &= 0,\ &(x,y) \in \mathcal{B}_0,\\
    u(x,y) = 0,\ v(x,y) &= 0,\ &(x,y) \in \mathcal{B}_1,
\end{align}
where $u(x,y)$ and $v(x,y)$ denote the x-component and y-component, respectively, of the velocity vector, while $p(x,y)$ is the pressure field. The domain $\Omega=(0,1)\times(0,1)$ represents a two-dimensional square cavity, where $\mathcal{B}_0$ denotes the top boundary and $\mathcal{B}_1$ represents the other three sides of the cavity. We set $\lambda_1=1$ and $\lambda_2=0.01$ following the setting in \cite{wang2021understanding}. We use $N_{r}={128}$ collocation points for the residual loss $\mathcal{L_\mathrm{r}}$ and 128 points for each boundary condition (resulting in a total of $N_{\mathrm{bc}}=512$ points). These training points are re-sampled for each iteration.

\paragraph{Flow past a circular cylinder.} In this case, we assume a two-dimensional spatial domain of $\Omega=[-15,25]\times[-8,8]$, where a circular cylinder with a diameter of $D=1$ is positioned at the origin. Following the configurations outlined in \cite{raissi2019physics}, we consider these boundary conditions: a uniform free stream velocity profile with $u_{\infty}=1$ is prescribed at the left boundary, a zero pressure outflow condition is enforced at the right boundary, situated 25 diameters downstream of the cylinder, and a periodic condition is imposed for the top and bottom boundaries. We assume a kinematic viscosity of $\nu=0.01$, resulting in a corresponding Reynolds number of $Re=u_{\infty}D/\nu=100$.  
The system is governed by the unsteady-state Navier-Stokes equations as 
\begin{align}
    u_t+\lambda_1(uu_x+vu_y)+p_x-\lambda_2(u_{xx}+u_{yy})&=0,\ &(x,y)\in\Omega, \ t\in[0,T]\\
    u_t+\lambda_1(uv_x+vv_y)+p_y-\lambda_2(v_{xx}+v_{yy})&=0,\ &(x,y)\in\Omega,\ t\in[0,T]\\ 
    u_x+v_y&=0, \ &(x,y)\in\Omega,\ t\in[0,T]
\end{align}
where $\lambda_1$ is set to 1, $\lambda_2=1/Re=0.01$ and $T=20$. We focus on the solution within the wake region behind the cylinder, which corresponds to the spatial domain denoted as $\Omega'=[1,8]\times[-2,2]$. We consider the inverse problem formulation of this case, where $\lambda_1$ and $\lambda_2$ are assumed to be unknown. Given a limited set of measurements of the velocity fields, denoted as $\{t_\mathrm{m}^i, x_\mathrm{m}^i, y_\mathrm{m}^i, u_\mathrm{m}^i, v_\mathrm{m}^i\}_i^{N_\mathrm{m}}$, the PINNs are optimized to accurately predict the dynamics of this system and estimate the values of $\lambda_1$ and $\lambda_2$. The objective function $\mathcal{L}(\boldsymbol{\theta})$ is composed of the residual loss $\mathcal{L}_{\mathrm{r}}$ in \cref{eq:L_r} and the data loss $\mathcal{L}_{\mathrm{m}}$
\begin{align}
    \mathcal{L}(\boldsymbol{\theta}) &= \mathcal{L}_{\mathrm{r}}(\boldsymbol{\theta})+\mathcal{L}_{\mathrm{m}}(\boldsymbol{\theta}),\\
    \mathcal{L}_{\mathrm{m}}(\boldsymbol{\theta}) &= \frac{1}{N_\mathrm{m}}\sum_{i=1}^{N_\mathrm{m}}(||u'(t_\mathrm{m}^i, x_\mathrm{m}^i, y_\mathrm{m}^i; \boldsymbol{\theta})-u_\mathrm{m}^i||_2^2+||v'(t_\mathrm{m}^i, x_\mathrm{m}^i, y_\mathrm{m}^i; \boldsymbol{\theta})-v_\mathrm{m}^i||_2^2).
\end{align}

\begin{table}[htbp]
\color{black}
  \caption{The experimental configurations for 2D problems.}
  \label{tab:config_2d}
  \centering
  \begin{tabular}{c|cc|cc|ccc}
    \toprule
    & \multicolumn{2}{c}{Network} & \multicolumn{2}{c}{Adam} & \multicolumn{3}{c}{Data}\\
    Problems    & Depth & Width & Iters & lr   & $N_\mathrm{m}$ & $N_\mathrm{bc}$ & $N_\mathrm{r}$ \\
    \midrule
    Flow in a lid-driven cavity & 6 & 64 & 80k & 1e-3   & - & 128 & 128 \\
    Flow past a circular cylinder & 6 & 64 & 40k & 1e-3  & 5000 & - & 5000 \\
    \bottomrule
  \end{tabular}
\color{black}
\end{table}

\subsubsection{Main results}
\begin{figure*}[htbp]
\subfigure[$\mathrm{GELU}$ ($L_2$ relative error: 5.90\%)]{
\begin{minipage}{0.95\textwidth}
    \centering
    \includegraphics[width=1.00\textwidth]{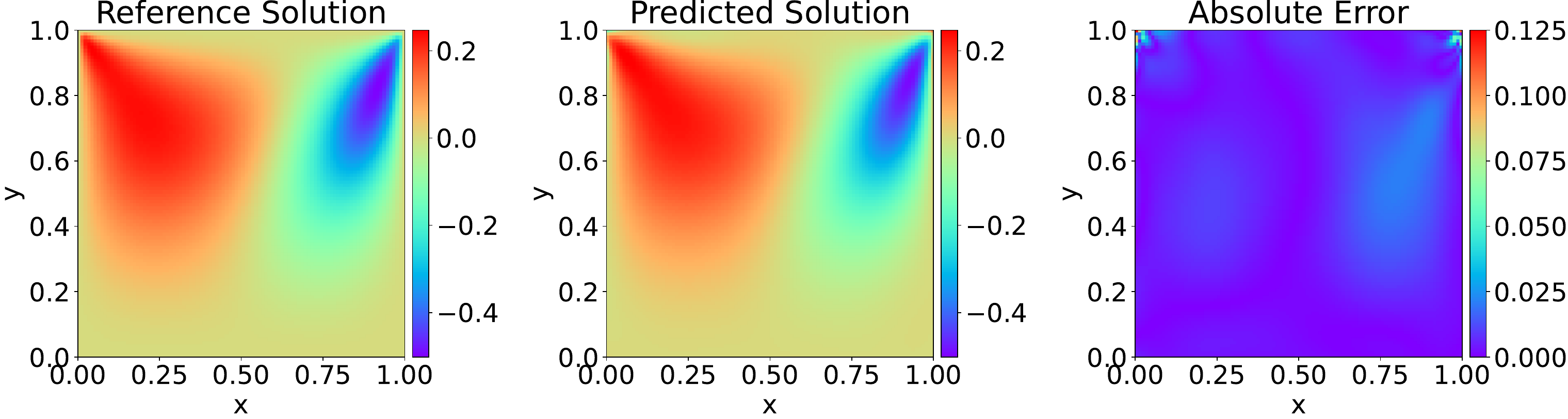}
\end{minipage}}
\subfigure[ACON ($L_2$ relative error: 6.68\%)]{
\begin{minipage}{0.95\textwidth}
    \centering
    \includegraphics[width=1.00\textwidth]{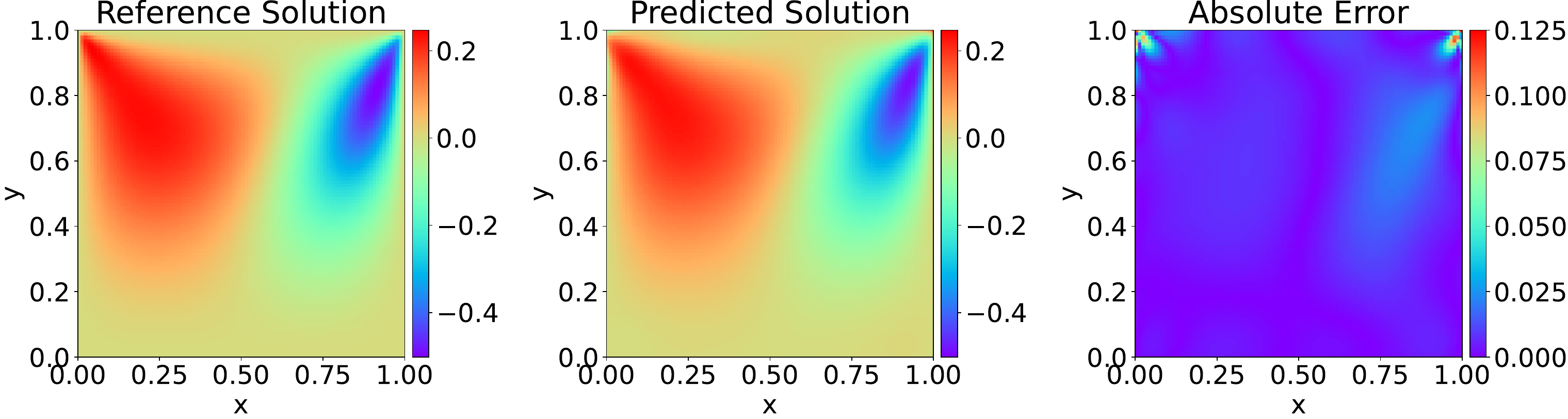}
\end{minipage}}
\subfigure[{\name} ($L_2$ relative error: 4.27\%)]{
\begin{minipage}{0.95\textwidth}
    \centering
    \includegraphics[width=1.00\textwidth]{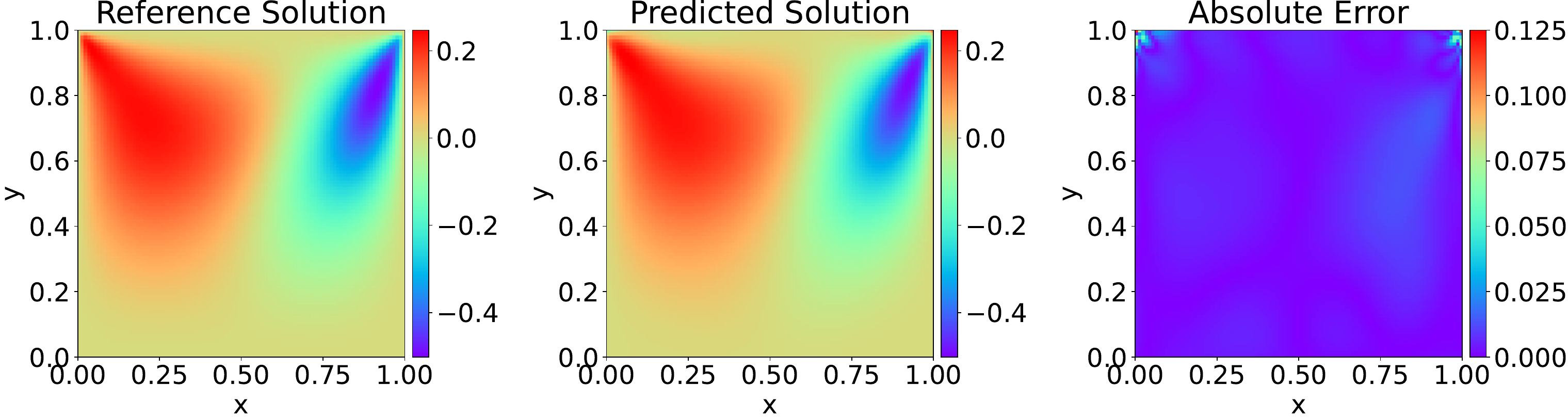}
\end{minipage}}
\subfigure[{\name} w/ AS ($L_2$ relative error: 3.73\%)]{
\begin{minipage}{0.95\textwidth}
    \centering
    \includegraphics[width=1.00\textwidth]{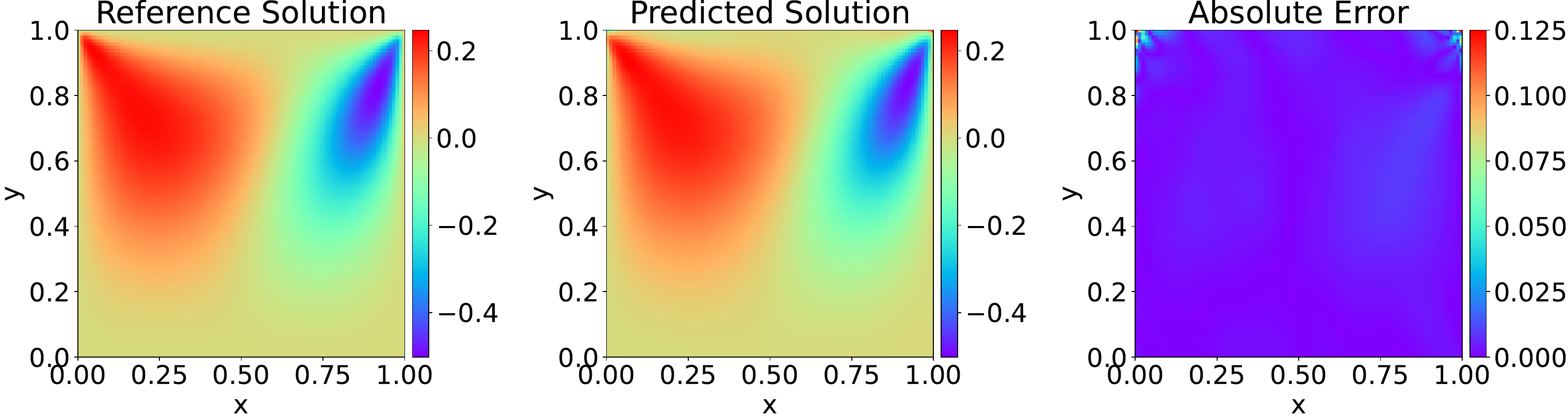}
\end{minipage}}
    \textcolor{black}{\caption{Flow in a lid-driven cavity, y-component of velocity $v$.}\label{fig:pred_detailed_flow_in_v}}
    
\end{figure*}

\begin{figure*}[ht]
\subfigure[$\mathrm{GELU}$ ($L_2$ relative error: 1.51\%)]{
\begin{minipage}{0.95\textwidth}
    \centering
    \includegraphics[width=1.00\textwidth]{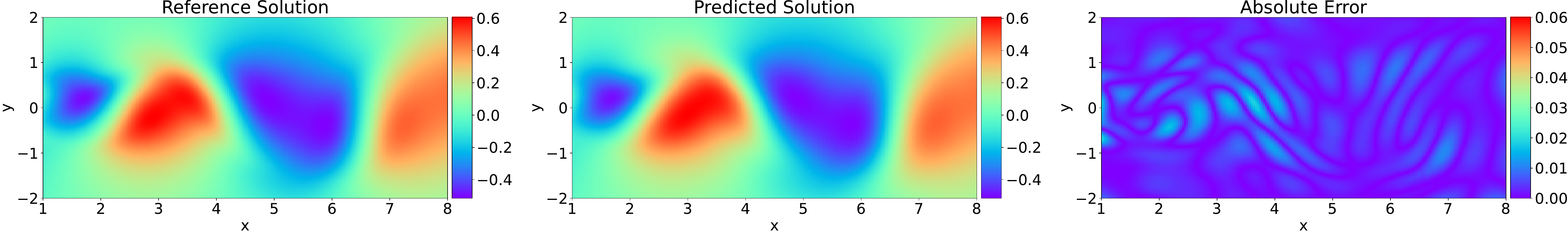}
\end{minipage}}
\subfigure[ACON ($L_2$ relative error: 2.40\%)]{
\begin{minipage}{0.95\textwidth}
    \centering
    \includegraphics[width=1.00\textwidth]{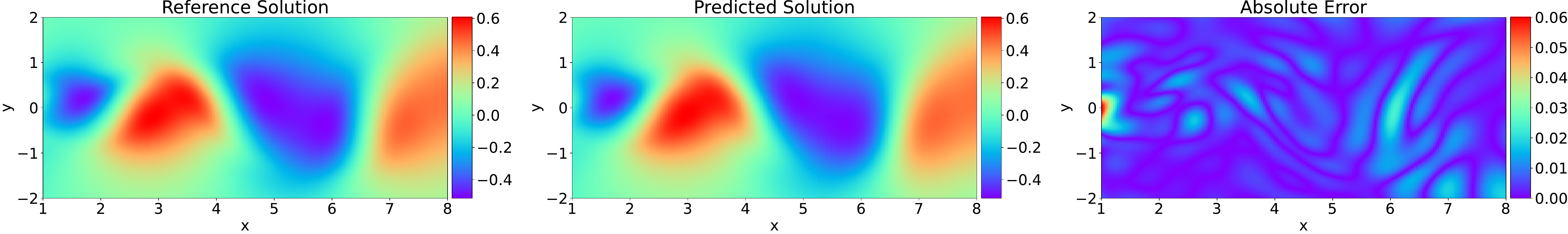}
\end{minipage}}
\subfigure[{\name} ($L_2$ relative error: 1.05\%)]{
\begin{minipage}{0.95\textwidth}
    \centering
    \includegraphics[width=1.00\textwidth]{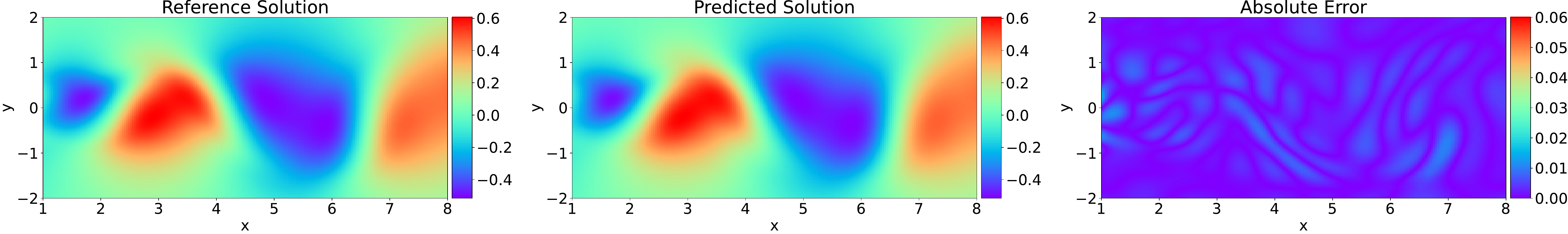}
\end{minipage}}
\subfigure[{\name} w/ AS ($L_2$ relative error: 0.70\%)]{
\begin{minipage}{0.95\textwidth}
    \centering
    \includegraphics[width=1.00\textwidth]{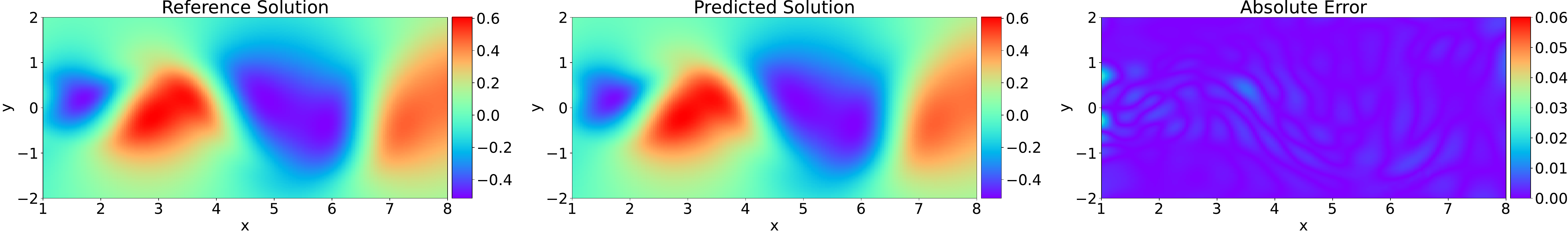}
\end{minipage}}
    \textcolor{black}{\caption{Flow past a circular cylinder, y-component of velocity $v$ (t=2.5).}\label{fig:pred_detailed_flow_past_v}}
    
\end{figure*}

We replicate the experimental settings outlined in \cref{1d_main_results}, maintaining consistency, with the exception that we exclusively employ the Adam optimizer during the training of PINNs. The detailed experimental configurations for 2D PDEs can be found in \cref{tab:config_2d}. The results are shown in  \cref{tab:error_ns_pde}. 
The proposed method exhibits superior performance across all evaluated metrics when compared to its candidate activation functions. In the scenario of flow in a lid-driven cavity, the utilization of {\name} yields a notable reduction in the error rates for both the x-component velocity $u$ and the y-component velocity $v$. Specifically, {\name} achieves a reduction of 11\% in the error rate for $u$ and 20\% in the error rate for $v$ when compared to the best-performing standard activation function, $\mathrm{GELU}$. In the case of flow past a circular cylinder, the performance of {\name} surpasses that of standard activation functions not only in predicting the dynamics of the system but also in estimating the values of $\lambda_1$ and $\lambda_2$. As a demonstration, {\name} achieves a significant performance improvement of 38\% and 27\% in the estimation of $\lambda_1$ and $\lambda_2$, respectively, compared to the $\mathrm{GELU}$ activation function. This superior performance of {\name} highlights its enhanced capability in handling inverse problems. Furthermore, it is worth noting that the performance of {\name} can be enhanced even further by incorporating the adaptive slope (AS) technique.  
We present the predictions of transverse velocity component $v$ for the lid-driven cavity flow in \cref{fig:pred_detailed_flow_in_v} and the flow past a circular cylinder in \cref{fig:pred_detailed_flow_past_v}. One can observe that the predicted solutions of {\name} exhibit a favorable consistency with respect to the reference solutions. More visualization results can be found in Appendix \cref{fig:pred_detailed_flow_in_u,fig:pred_detailed_flow_past_u,fig:pred_detailed_flow_past_p}.

\color{black}

\begin{table}
\color{black}
\centering
\caption{Experimental results on 2D PDEs. We report the $L_2$ relative error (\%) for each variable. In the case of flow past a circular cylinder, we present the averaged results obtained from 200 temporal snapshots covering the time interval from t=0 to t=20, with a time increment of 0.1. The best results are \textbf{bold-faced}, while the second best results are \underline{underlined}.}
\label{tab:error_ns_pde}
\resizebox{1.0\linewidth}{!}
 {
\begin{tabular}{c|cc|ccccc} 
\toprule
& \multicolumn{2}{c}{Flow in a lid-driven cavity} &  \multicolumn{5}{c}{Flow past a circular cylinder} \\
 Method & $u$ &  $v$& $u$ &  $v$ & $p$ & $\lambda_1$ & $\lambda_2$  \\
\midrule
$\mathrm{sin}$ & $5.36\pm0.95$ & $7.65\pm2.08$& $0.69\pm0.09$ & $1.97\pm0.30$& $2.59\pm0.39$ &  $0.15\pm0.03$ & $10.44\pm1.12$ \\
$\mathrm{tanh}$ & $7.65\pm1.18$ & $11.78\pm1.98$& $0.54\pm0.03$ & $1.70\pm0.06$& $2.50\pm0.13$ &  $0.09\pm0.01$ & $7.53\pm0.52$\\
$\mathrm{GELU}$ & $4.48\pm0.10$ & $5.71\pm0.11$& $0.52\pm0.02$ & $1.45\pm0.04$ & $2.24\pm0.05$ & $0.08\pm0.03$ & $6.83\pm0.66$  \\
$\mathrm{Swish}$ & $4.90\pm0.49$ & $7.02\pm1.15$& $0.61\pm0.05$ & $1.68\pm0.16$ & $2.74\pm0.08$ & $1.17\pm0.04$ & $9.29\pm1.29$ \\
$\mathrm{Softplus}$& $8.78\pm0.43$ & $14.63\pm1.15$& $1.61\pm0.12$ & $4.52\pm0.49$ & $6.76\pm0.77$ & $0.78\pm0.10$ & $11.76\pm0.24$ \\
\midrule
SLAF & $4.78\pm0.09$ & $6.42\pm0.32$& $1.39\pm0.64$ & $3.20\pm1.47$ & $4.13\pm1.21$ & $0.34\pm0.22$ & $13.70\pm3.01$ \\
PAU & $87.43\pm8.78$ & $97.36\pm3.12$& $1.24\pm0.05$ & $3.34\pm0.12$ & $7.56\pm0.47$ & $1.11\pm0.07$ & $19.27\pm8.00$ \\
ACON & $4.67\pm0.17$ & $6.27\pm0.41$& $0.70\pm0.13$ & $1.87\pm0.36$ & $3.04\pm0.56$ & $0.10\pm0.03$ & $6.52\pm1.79$ \\
{\textbf{\name}} & $\underline{3.99\pm0.07}$ & $\underline{4.55\pm0.28}$&$\underline{0.48\pm0.01}$ &$\underline{1.35\pm0.05}$ & $\underline{2.11\pm0.02}$ & $\underline{0.05\pm0.02}$ & $\underline{4.97\pm0.13}$ \\
{\textbf{\name} (AS)}& $\bm{3.59\pm0.09}$ & $\bm{3.97\pm0.24}$& $\bm{0.38\pm0.01}$ & $\bm{1.07\pm0.01}$ & $\bm{1.87\pm0.06}$ & $\bm{0.02\pm0.006}$ & $\bm{4.58\pm0.11}$\\
\bottomrule
\end{tabular}
}
\color{black}
\end{table}

\subsection{Ablation study}
\label{ablation}
We ablate different design choices to provide a better understanding of the proposed method. For all ablation experiments, we deploy the layer-wise {\name} to solve the convection equation, unless explicitly stated otherwise. The adaptive slope is deactivated for a clear comparison.

\paragraph{Candidate function set.} Here we study the influence of candidate set by gradually increasing its size as shown in \cref{tab:ablaton_function_set}. We begin with the two most commonly-used activation functions, the $\mathrm{sin}$ and $\mathrm{\tanh}$ functions. 
In this case, {\name} is shown to achieve competitive result compared with its two candidate functions. 
Based on that, we observe that adding new activation functions could lead to a better performance. Surprisingly, the introduction of $\mathrm{GELU}$ can still improve the performance despite its poor results on this problem. We argue that this performance gain arises from the enlarged search space. 
We notice that the performance tends to saturate as the number of candidate functions increases. This indicates that {\name} has a good robustness to the choice of candidate function set if this set contains activation functions with sufficient diversity. Notably, the additional parameters of {\name} is negligible compared with total number of weights. 

\begin{table}
\centering
\caption{{\name} ablation experiments on the convection equation. Comparisons of different candidate function set. We repeat each experiments 5 times and report the average $L_2$ relative error (\%). Default settings are marked in \colorbox{gray!20}{gray}.}

\label{tab:ablaton_function_set}
\begin{tabular}{lcc} 
\toprule
Function set $\mathcal{F}$ & \#params & error (\%) \\
\midrule
\{$\mathrm{sin}$, $\mathrm{tanh}$, $\mathrm{GELU}$, $\mathrm{Swish}$, $\mathrm{Softplus}$\} & 16922 & \colorbox{gray!20}{$\bm{0.06\pm0.03}$} \\
  \{$\mathrm{sin}$, $\mathrm{tanh}$, $\mathrm{GELU}$, $\mathrm{Swish}$\} & 16917 & $\bm{0.06\pm0.03}$ \\
   \{$\mathrm{sin}$, $\mathrm{tanh}$, $\mathrm{GELU}$\}  & 16912 & $0.10\pm0.12$\\
   \{$\mathrm{sin}$, $\mathrm{tanh}$\} & 16907 & $0.33\pm0.04$\\
 \midrule
$\mathrm{sin}$ & 16897 & $0.36\pm0.15$\\
  $\mathrm{tanh}$ & 16897 &  $6.83\pm4.79$\\
  $\mathrm{GELU}$ & 16897 &  $39.29\pm35.51$\\
\bottomrule
\end{tabular}
\end{table}

\paragraph{Gate functions.} We also compare the softmax function with other gate functions in \cref{tab:ablaton_search_space}. The identity function $G(\alpha_i)=\alpha_i$ includes all linear combinations of candidate activation functions into the search space; while the sigmoid function $G(\alpha_i) = 1/(1+\mathrm{exp}(-\alpha_i))$ restricts the coefficients to be between 0 and 1. We also consider the $L_1$-normalization  $G(\alpha_i) = \alpha_i/\sum_{j=1}^N |\alpha_j|$, which allows for negative coefficients and keeps the competition between candidate functions. The trainable $\{\alpha_i\}_{i=1}^N$ are initialized as zeros for sigmoid and softmax functions and as $1/N$ for identity and $L_1$-normalization functions, where $N$ denotes the number of candidate functions. One can observe that the results of softmax function are more stable and more accurate. This implies proper restrictions of the search space can work as a regularization to the learning of {\name}.

\paragraph{Learnable coefficients of {\name}.} To demonstrate that the learnable coefficients of {\name} is effective and necessary, we compare {\name} to two variants with fixed coefficients. The first one initializes its parameters as zeros and the second one initializes its parameters from a standard normal distribution. As shown in \cref{tab:ablaton_search_space}, the automatically learned coefficients outperforms evenly distributed coefficients and random sampled coefficients by a large margin. One can also observe that the performance gap between the random and constant initialization is eliminated if these parameters are trainable, which demonstrates the effectiveness of {\name}'s optimization. 

\begin{table}
\centering
\caption{{\name} ablation experiments on the convection equation. Comparisons of different search spaces. We repeat each experiments 5 times and report the average $L_2$ relative error (\%). Default settings are marked in \colorbox{gray!20}{gray}.}

\label{tab:ablaton_search_space}
\begin{tabular}{cccc} 
\toprule
$G(\cdot)$ & Init & Learnable & error (\%) \\
\midrule
          identity & 1/N & \checkmark & $0.22\pm0.22$ \\
          $L_1$-norm & 1/N & \checkmark & $1.18\pm0.93$ \\
          sigmoid & Zero & \checkmark & $0.89\pm0.83$ \\
          softmax & Zero & \checkmark & \colorbox{gray!20}{$\bm{0.06\pm0.03}$} \\
 \midrule
         softmax & Zero & \XSolidBrush & $1.60\pm1.12$ \\
         softmax & Random & \XSolidBrush & $0.32\pm0.23$ \\
         softmax & Random & \checkmark & $0.09\pm0.05$ \\

\bottomrule
\end{tabular}
\end{table}

\textcolor{black}{\paragraph{Network architectures.} We validate the performance of our method across a range of network depths and widths. The results are summarized in \cref{tab:ablaton_network_size}. Remarkably, the proposed {\name} consistently achieves the best performance across all investigated scenarios, surpassing standard activation functions and other adaptive activation functions. The empirical evidence, as depicted in  \cref{fig:train_loss_diff_network}, demonstrates that the training losses of {\name} exhibit rapid and stable convergence across varying network depths and widths, affirming the robustness of our method in accommodating diverse network architectures.}
 
\begin{table}
\color{black}
\centering
\caption{{\name} ablation experiments on the Allen-Cahn equation. Comparisons of different network depths and widths. We repeat each experiments 5 times and report the average $L_2$ relative error (\%). Default settings are marked in \colorbox{gray!20}{gray}.}
\label{tab:ablaton_network_size}
\resizebox{1.0\linewidth}{!}
 {
\begin{tabular}{c|ccc|ccc} 
\toprule
 & \multicolumn{3}{c}{Depth=4} & \multicolumn{3}{c}{Width=32} \\
 Method & \colorbox{gray!20}{Width=32} &  Width=64 & Width=96 & \colorbox{gray!20}{Depth=4} & Depth=6 & Depth=8 \\
\midrule
$\mathrm{sin}$ & $3.57\pm0.64$ & $3.62\pm0.69$ & $3.56\pm0.65$ & $3.57\pm0.64$ & $0.75\pm0.48$ & $0.42\pm0.25$ \\
$\mathrm{tanh}$ & $1.34\pm0.54$ & $0.99\pm0.26$ & $0.72\pm0.28$ & $1.34\pm0.54$ & $0.68\pm0.47$ & $0.39\pm0.12$ \\
 \midrule
SLAF & $33.93\pm9.97$ & $32.56\pm12.51$ & $28.04\pm5.14$ & $33.93\pm9.97$ & $8.55\pm1.66$ & $3.30\pm0.92$ \\
PAU & $43.83\pm15.18$ & $33.61\pm13.86$ & $26.93\pm16.40$ & $43.83\pm15.18$ & $21.25\pm19.14$ & $32.00\pm21.90$ \\
ACON & $3.88\pm1.82$ & $3.08\pm1.16$ & $2.31\pm0.44$ & $3.88\pm1.82$ & $3.34\pm2.42$ & $2.46\pm2.74$ \\
{\textbf{\name}} & $\bm{0.76\pm0.26} $ & $\bm{0.66\pm0.04}$ & $\bm{0.45\pm0.07}$ & $\bm{0.76\pm0.26}$ & $\bm{0.26\pm0.01}$ & $\bm{0.22\pm0.01}$ \\
\bottomrule
\end{tabular}}
\color{black}
\end{table}

\subsection{Understanding {\name} through the lens of neural tangent kernel}
\label{exp_ntk}
\textcolor{black}{The neural tangent kernel is proposed to describe the evolution of neural networks under the assumptions of infinite network width and infinitesimal learning rate}~\cite{jacot2018neural}. Its eigenvalues can be leveraged to analyse the rate of convergence as shown in previous works~\cite{jacot2018neural, tancik2020fourier, wang2022and}.  We show that the introduction of {\name} makes the NTK learnable. Empirically, we observe the optimization of {\name} adapts the NTK's eigenvalue spectrum to the underlying PDE system and leads to an improvement on the 
average eigenvalue, which partially explains the stable and fast convergence of {\name}. 

\paragraph{Brief introduction of NTK.} 
The neural tangent kernel of a $L$-layer fully-connected network $u^{(L)}$ with parameters $\boldsymbol{\theta}$ is defined as 
\begin{equation}
    \Theta^{(L)}(\vx,\vx')=\langle \frac{\partial u^{(L)}(\vx|\boldsymbol{\theta})}{\partial \boldsymbol{\theta}}, \frac{\partial u^{(L)}(\vx'|\boldsymbol{\theta})}{\partial \boldsymbol{\theta}} \rangle=\sum_{\theta_i}\frac{\partial u^{(L)}(\vx|\boldsymbol{\theta})}{\partial \theta_i} \frac{\partial u^{(L)}(\vx|\boldsymbol{\theta})}{\partial \theta_i}
\end{equation}
where $\vx$ and $\vx'$ are two inputs. As demonstrated in ~\cite{jacot2018neural}, in the infinite-width limit and under suitable conditions, the NTK $\Theta^{(L)}$ converges to a deterministic kernel $\Theta^{(L)}_{\infty}$ at initialization and stays constant during the training. This limiting kernel can be represented as 
\begin{align}
      \Theta^{(L)}_{\infty} &= \sum_{l=1}^{L} \bigg(\Sigma^{(l)}(\vx,\vx')\prod_{l'=l+1}^L\dot{\Sigma}^{(l')}(\vx,\vx')\bigg), \label{eq:limiting_ntk} \\
      \Sigma^{(l+1)}(\vx, \vx') &= \mathbb{E}_{f\sim \mathcal{N}(0, \Sigma^{(l)})}[\sigma(f(\vx))\sigma(f(\vx'))] + \beta^2, \label{eq:cor_sigma} \\
    \dot{\Sigma}^{(l+1)}(\vx, \vx') &= \mathbb{E}_{f\sim \mathcal{N}(0, \Sigma^{(l)})}[\dot{\sigma}(f(\vx))\dot{\sigma}(f(\vx'))] + \beta^2 \label{eq:cor_hat_sigma} . 
\end{align}
Note $\Sigma^{(1)}(\vx, \vx') = \frac{1}{n_0}\vx^T\vx' + \beta^2$. The expectations in \cref{eq:cor_sigma,eq:cor_hat_sigma} are taken with respect to a centered Gaussian process $f$ with covariance $\Sigma^{(l)}$; $n_0$ is the dimension of input $\vx$ and $\beta$ is a scaling factor; $\sigma$ and $\dot{\sigma}$ denote the activation function and its derivative, respectively. Note that the covariance $\Sigma^{(l)}$ depends on the choice of activation functions and so dose the NTK $\Theta^{(L)}_{\infty}$.

\paragraph{{\name} and the learnable NTK.} For simplicity, we derive the NTK of standard neural networks with {\name}. The NTK of PINNs with {\name} can be derived in a similar way. 
Firstly, we assume that the optimization of neural networks with {\name} can be decomposed into two phases, where we learn the coefficients of {\name} in the first phase and then train the parameters of neural network in the second phase.
This assumption is reasonable as the number of parameters of {\name} is far less than those of networks and they quickly converge at the early stage of training. Empirically, we observe that a short warmup (2.5\% of the whole schedule) is sufficient for {\name} to learn suitable activation functions and to achieve competitive performance compared with the counterpart whose coefficients are updated during the whole schedule (0.58\% vs. 0.60\% on the Allen-Cahn equation). By decoupling the updates of {\name} and networks, we find the second optimization phase is equivalent to the training of a standard network with a learned activation function, whose NTK is derived as 
\begin{align}
     \bar{\Theta}^{(L)}_{\infty} = \sum_{l=1}^{L} \bigg((\sum_{i=1}^N\sum_{j=1}^N G(\alpha_i)G(\alpha_j)\Sigma^{(l)}_{\sigma_i\sigma_j}(\vx,\vx'))&\prod_{l'=l+1}^L(\sum_{i=1}^N\sum_{j=1}^N G(\alpha_i)G(\alpha_j)\dot{\Sigma}^{(l')}_{\sigma_i\sigma_j}(\vx,\vx'))\bigg), \label{eq:piac_ntk} \\
      \Sigma^{(l+1)}_{\sigma_i\sigma_j}(\vx, \vx') &= \mathbb{E}_{f\sim \mathcal{N}(0, \Sigma^{(l)})}[\sigma_i(f(\vx))\sigma_j(f(\vx'))], \\
    \dot{\Sigma}^{(l+1)}_{\sigma_i\sigma_j}(\vx, \vx') &= \mathbb{E}_{f\sim \mathcal{N}(0, \Sigma^{(l)})}[\dot{\sigma}_i(f(\vx))\dot{\sigma}_j(f(\vx'))]. 
\end{align}
We omit the factor $\beta$ for convenience. One can observe that the covariance $\Sigma^{(l)}$ in \cref{eq:limiting_ntk} is replaced with a weighted sum of $\Sigma^{(l)}_{\sigma_i\sigma_j}$, which is calculated with different combinations of candidate activation functions. Note that the weights are learned in the first training phase. To conclude, the introduction of {\name} leads to a learnable NTK which could be adapted to the underlying PDE. 

\begin{figure*}[h]
\subfigure[Candidate functions]{
\begin{minipage}{.47\textwidth}
    \centering
    \includegraphics[width=0.9\textwidth]{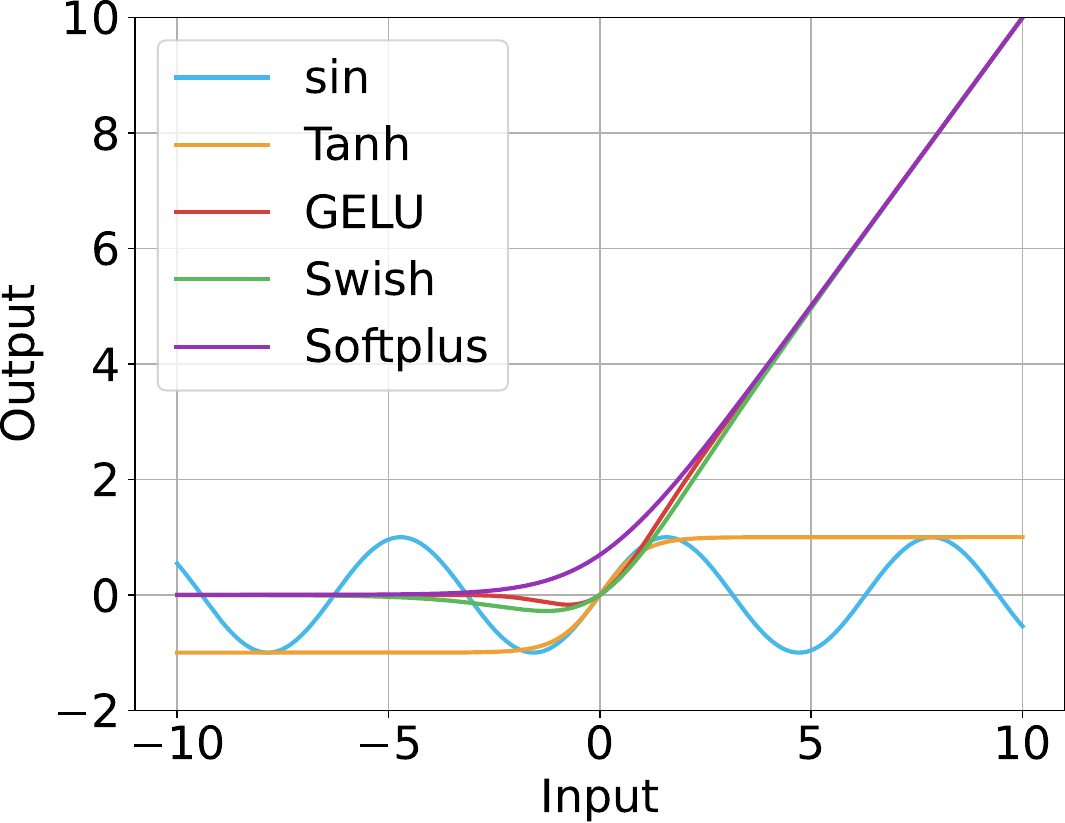}
\end{minipage}}
\subfigure[Convection equation]{
\begin{minipage}{.47\textwidth}
    \centering
    \includegraphics[width=0.9\textwidth]{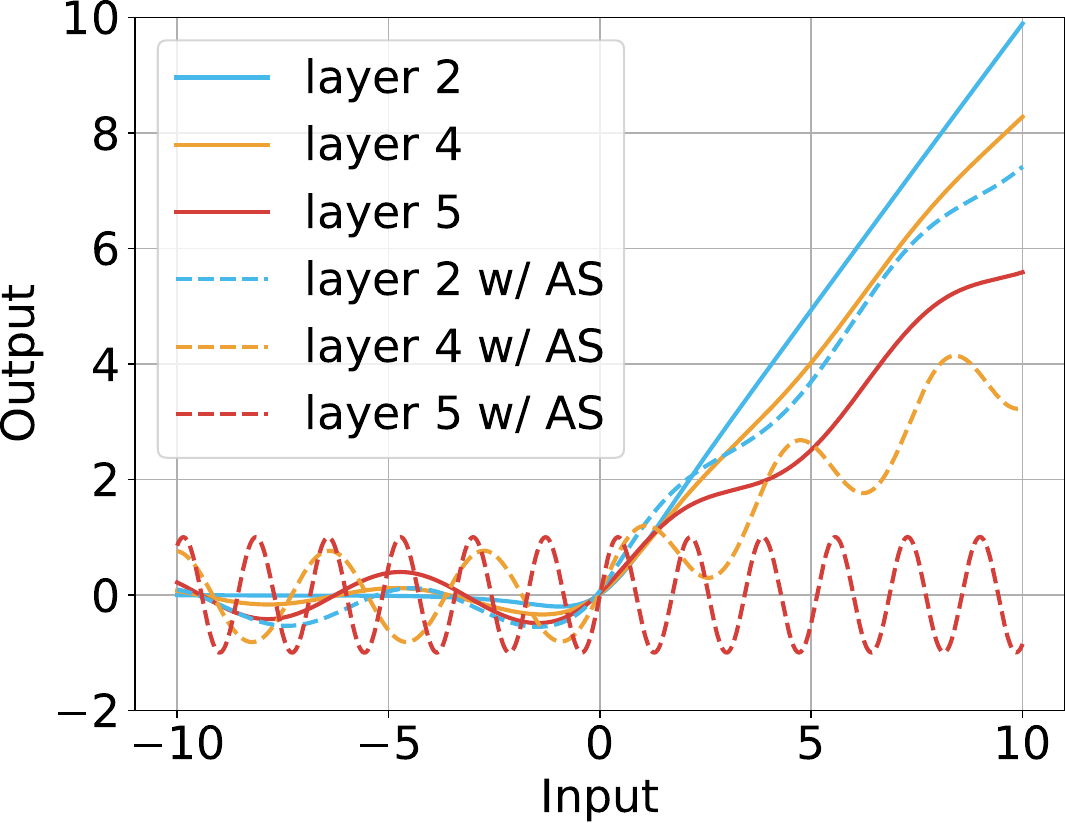}
\end{minipage}
}
\\
\subfigure[KdV equation]{
\begin{minipage}{.47\textwidth}
    \centering
    \includegraphics[width=0.9\textwidth]{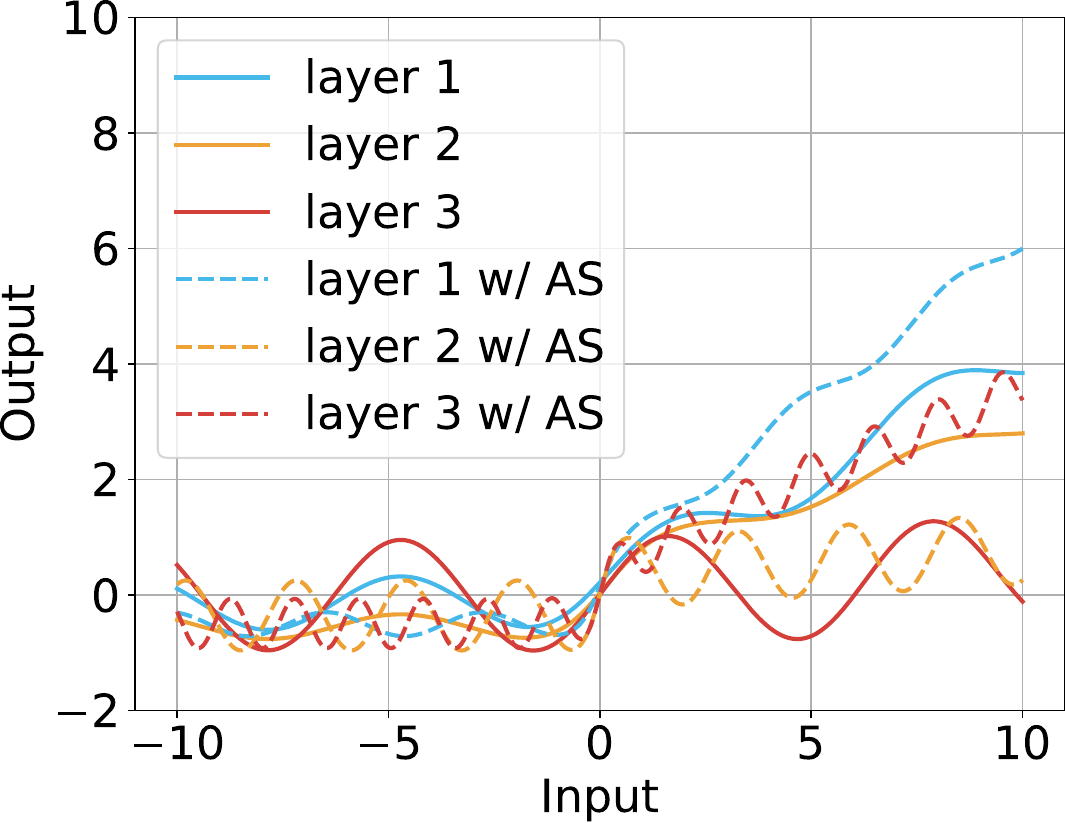}
\end{minipage}}\hspace{1.5mm}\subfigure[Eigenvalue spectrum]{\begin{minipage}{.47\textwidth}
    \centering
    \includegraphics[width=0.9\textwidth]{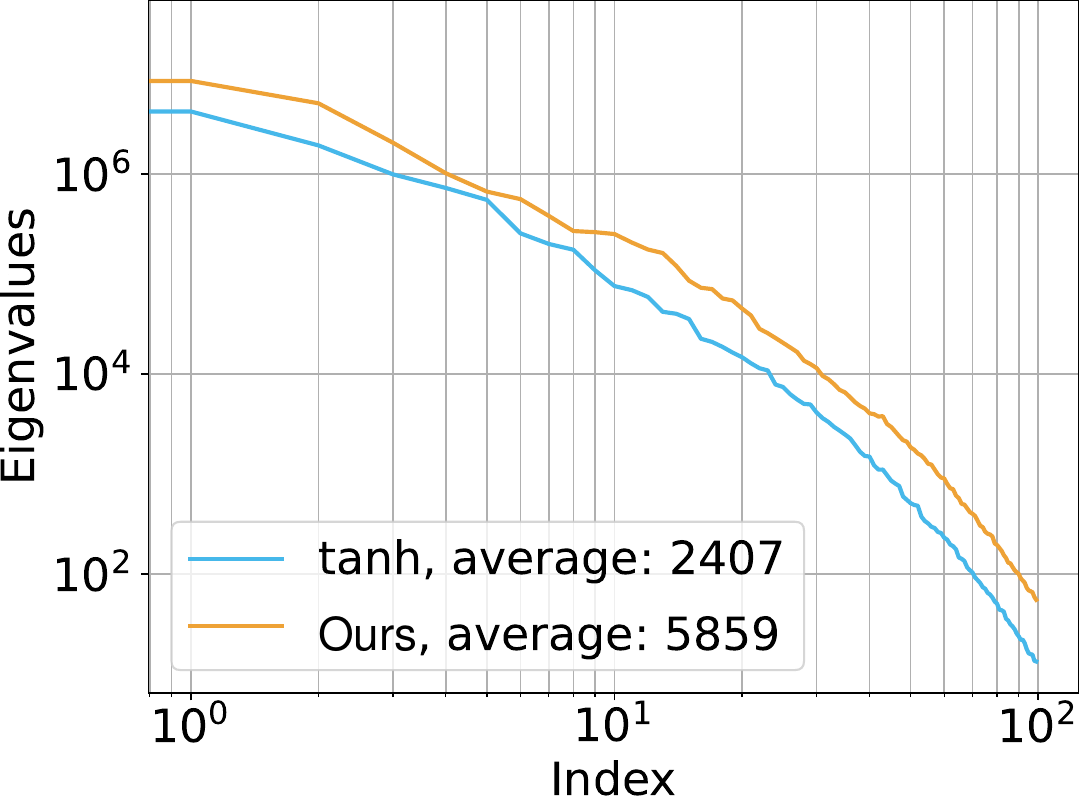}
    \label{fig:eig_spec}
\end{minipage}
}
    \caption{Visualization of the learned activation functions and the eigenvalue spectrum. (a) Candidate functions of {\name}. (bc) The learned activation functions for the convection equation and the KdV equation. The curves of {\name} with and without the adaptive slope are presented. (d) The eigenvalue spectrum of NTK matrix on the training data of the Allen-Cahn equation. The word "Ours" refers to ABU-PINN method.}
    \label{fig:learned_piac}
\end{figure*}

\paragraph{The eigenvalue spectrum of NTK with {\name}.} Through the optimization of {\name}, we can modify the limiting NTK $\bar{\Theta}^{(L)}_{\infty}$, which corresponds to a change in the convergence behavior of the neural network. As revealed by previous works~\cite{wang2022and,wang2022respecting}, the NTK's eigenvalues determine the convergence rate of the training loss. We then show empirically that the optimization of {\name} has an effect on the NTK's eigenvalue spectrum. As we can see in \cref{fig:eig_spec}, {\name} leads to a greater average eigenvalue compared with the best standard activation function on the Allen-Cahn equation (5859 vs. 2407), which 
implies a faster convergence rate.

\section{Summary and discussion}
\label{conclusion}

Despite the recent success, PINNs suffer from the ill-conditioned optimization.  
In this paper, we shed light on the relationship between activation functions and the optimization difficulty of PINNs. We reveal the high sensitivity of PINNs to the choice of activation functions and relate it to various characteristics of the underlying PDE system.  
To avoid the inefficient manual selection of activation functions, we propose to learn specialized activation functions automatically for PINNs to solve different problems. We compare different designs of adaptive activation functions and discuss their limitations in the context of PINNs. To remove obstacles to apply adaptive activation functions to PINNs, the proposed ABU-PINN tailors the idea of learning combinations of activation functions to the optimization of PDE-based constraints. Extensive experiments on a series of challenging benchmarks demonstrate the effectiveness of the proposed methods. 
In summary, our work presents a practical solution to the optimization difficulty from the perspective of activation functions. 
We hope our work could provide new insights and inspire further study on the convergence issue of PINNs from the point of view of network architecture design.

The limitations of our work are two-fold. The first limitation is that we only consider integer-order PDEs in the experiments, while different types of PDEs could have different effects on the training of adaptive activation functions. We leave the further exploration on fractional PDEs and stochastic PDEs for the future work. Another limitation is that we only focus on the empirical evaluation of adaptive activation functions, while theoretical analysis of the impact of PDE-based constraints on different algorithms is limited. We regard it as a promising direction to provide a rigorous theoretical analysis to gain insights into the activation function design for PINNs, and we will explore it in the future work.

\section*{Acknowledgments}
The work is supported in part by the National Natural Science Foundation of China under Grants 62276150 and the Guoqiang Institute of Tsinghua University.

\appendix

\section{Standard activation functions}
\label{sec:standard_act}
We compare several standard activation functions for solving a series of PDEs. The details are shown as follows. 

\begin{itemize}
    \item $\mathrm{sin}$:
    \begin{equation}
        f(x) = \mathrm{sin}(\beta*x),
    \end{equation}
    where scaling factor $\beta$ is set to 1 as default. 
    \item $\mathrm{cos}$:
    \begin{equation}
        f(x) = \mathrm{cos}(\beta*x),
    \end{equation}
     where $\beta$ is set to 1.
    \item $\mathrm{exp}$:
    \begin{equation}
        f(x) = e^{(\beta*x)}-1,
    \end{equation}
     where $\beta$ is set to 1 as default. For the \textbf{P1} setting of 1D Poisson's equation, $\beta$ is set to 0.25 for better performance.
     \item $\mathrm{tanh}$:
    \begin{equation}
        f(x) = \frac{e^x-e^{-x}}{e^x+e^{-x}}.
    \end{equation}
    \item $\mathrm{sigmoid}$:
    \begin{equation}
        f(x) = \frac{1}{1+e^{-x}}.
    \end{equation}
    \item $\mathrm{Softplus}$~\cite{dugas2000incorporating,glorot2011deep}:
    \begin{equation}
        f(x) = \frac{1}{\beta}\mathrm{log}(1+e^{\beta*x}),
    \end{equation}
    where $\beta=1$. $\mathrm{Softplus}$ can be regard as a smooth version of ReLU. 
    \item the Exponential Linear Unit ($\mathrm{ELU}$)~\cite{clevert2015fast}:
    \begin{equation}
        f(x)=\left\{
        \begin{aligned}
        &x, & \mathrm{if}\ x \geq 0, \\
        & e^x-1, & \mathrm{if}\ x < 0.
        \end{aligned}
        \right.
    \end{equation}
    \item the Gaussian Error Linear Unit ($\mathrm{GELU}$)~\cite{hendrycks2016bridging}:
    \begin{equation}
        f(x) = x * \Phi(x),
    \end{equation}
    where $\Phi(x)$ is the cumulative distribution function of the standard Gaussian distribution.
    \item $\mathrm{Swish}$~\cite{ramachandran2017searching}:
    \begin{equation}
        f(x) = \frac{x}{1+e^{-x}}.
    \end{equation}
\end{itemize}

\section{Burgers' equation with varying viscosity}
In this section, we conduct a comparative analysis of the performance obtained by different activation functions on Burgers' equation, considering varying viscosity values. As the viscosity approaches very small values, the smooth viscous solutions exhibit non-uniform convergence towards the appropriate discontinuous shock wave. This phenomenon presents a challenging problem for PINNs due to the unique complexities associated with capturing discontinuous features accurately~\cite{fuks2020limitations}.

\begin{figure*}[ht]

\centering
\includegraphics[width=0.5\textwidth]{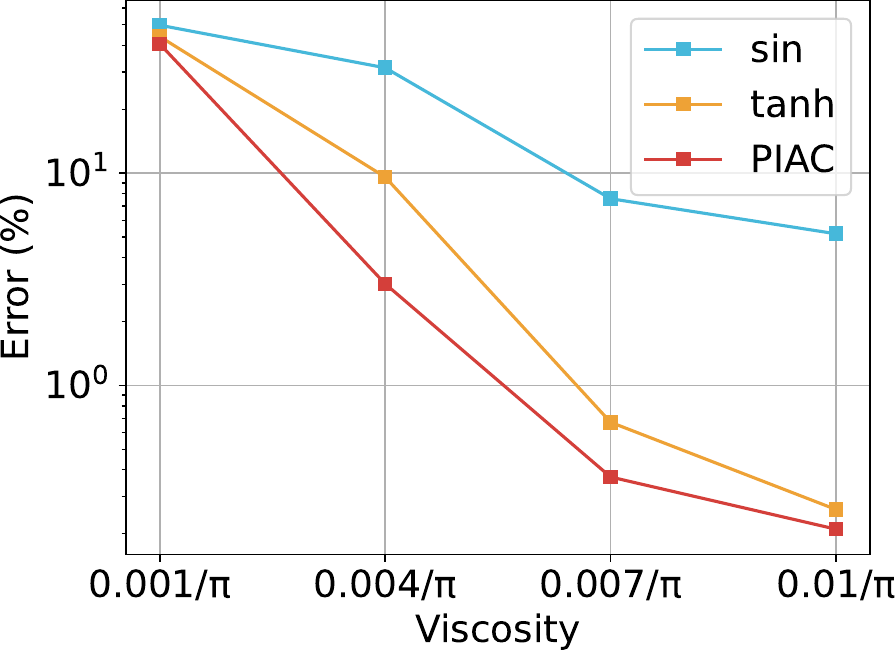}
\textcolor{black}{\caption{The performance on Burgers' equation with varying viscosity. We repeat each experiments 3 times and report the average $L_2$ relative error (\%) }\label{fig:burger_range_results}}
\end{figure*}

\begin{figure*}[htbp]
\subfigure[Reference solution]{
\begin{minipage}{0.95\textwidth}
    \centering
    \includegraphics[width=1.00\textwidth]{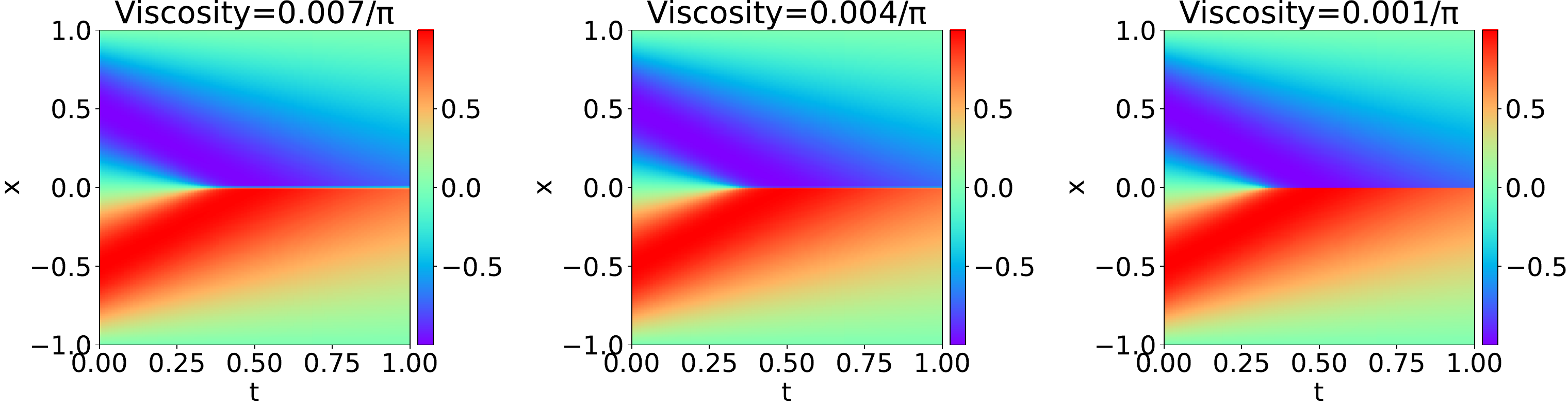}
\end{minipage}}
\subfigure[$\mathrm{sin}$]{
\begin{minipage}{0.95\textwidth}
    \centering
    \includegraphics[width=1.00\textwidth]{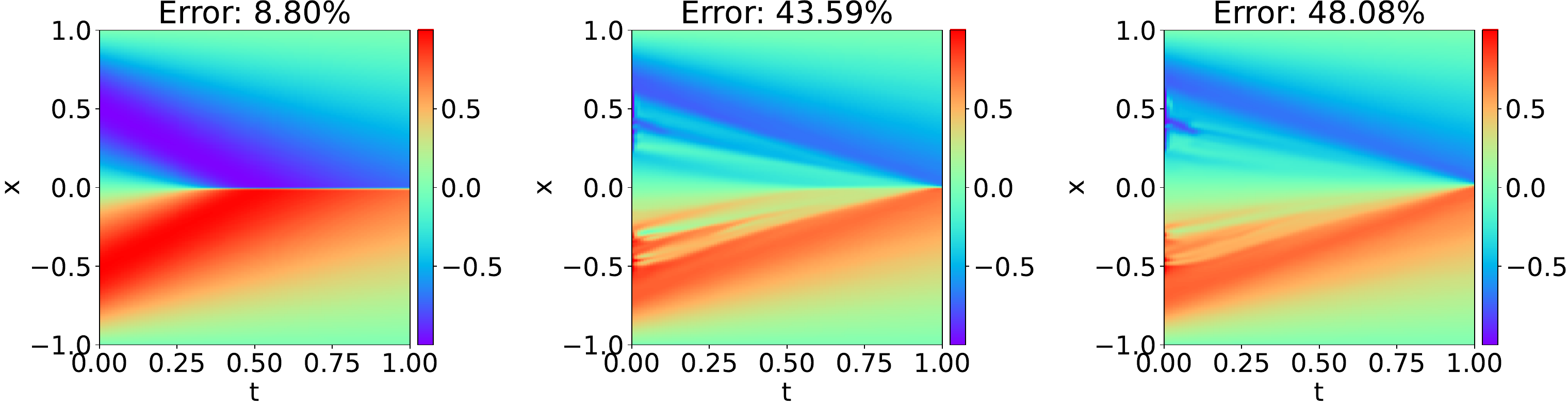}
\end{minipage}}
\subfigure[$\mathrm{tanh}$]{
\begin{minipage}{0.95\textwidth}
    \centering
    \includegraphics[width=1.00\textwidth]{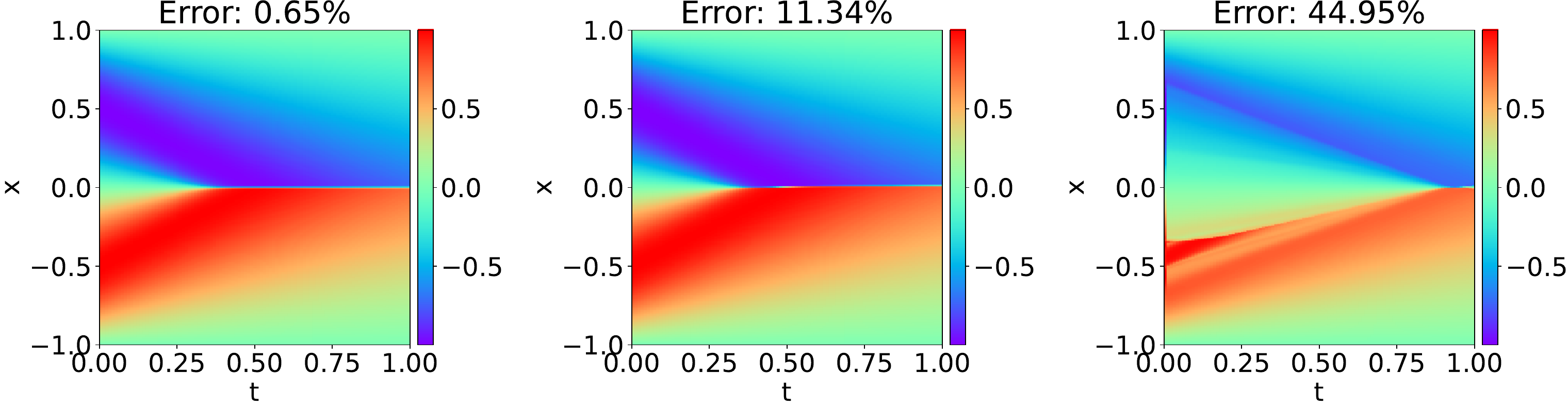}
\end{minipage}}
\subfigure[{\name}]{
\begin{minipage}{0.95\textwidth}
    \centering
    \includegraphics[width=1.00\textwidth]{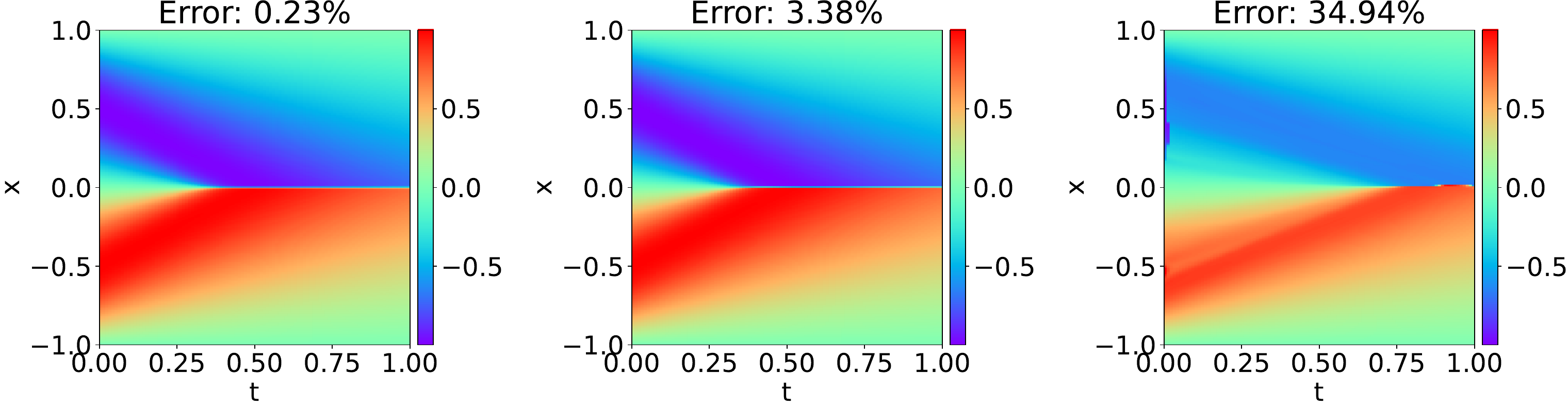}
\end{minipage}}
    \textcolor{black}{\caption{Predicted solutions for Burgers' equation with viscosity values of $0.007/\pi$ (left), $0.004/\pi$ (middle) and $0.001/\pi$ (right).}\label{fig:pred_detailed_burger_range}}
\end{figure*}

Starting from the case in \cref{exp_1d_time_dependent} with a viscosity of $0.01/\pi$, we investigate three additional cases with viscosity values of \{$0.007/\pi$, $0.004/\pi$, $0.001/\pi$\}. The same training configuration, as outlined in \cref{1d_main_results}, is utilized for all cases. The number of training iteration with Adam optimization is set to 15000 for the case with a viscosity of $0.007/\pi$ and increased to 40000 for the other two cases. The results are summarized in \cref{fig:burger_range_results} and \cref{fig:pred_detailed_burger_range}. Notably, the proposed method consistently outperforms standard activation functions as the viscosity value is decreased, showcasing its superior efficacy. As a prominent example, the proposed {\name} successfully achieves accurate predictions for the case with a viscosity of $0.004/\pi$. In contrast, the $\mathrm{sin}$ activation function fails to produce satisfactory results, while the $\mathrm{tanh}$ activation function yields distorted predictions around the time of 0.5. However, when the viscosity is further decreased to $0.001/\pi$,  neither standard activation functions nor {\name} can obtain satisfactory predictions. A plausible explanation for this observation is that the continuous nature of the proposed method may restrict its capability to enhance the current framework of PINNs in effectively capturing discontinuities.

\section{Visualization results} 
\begin{figure*}[htbp]
\subfigure{
\begin{minipage}{0.98\textwidth}
    \centering
    \includegraphics[width=1.00\textwidth]{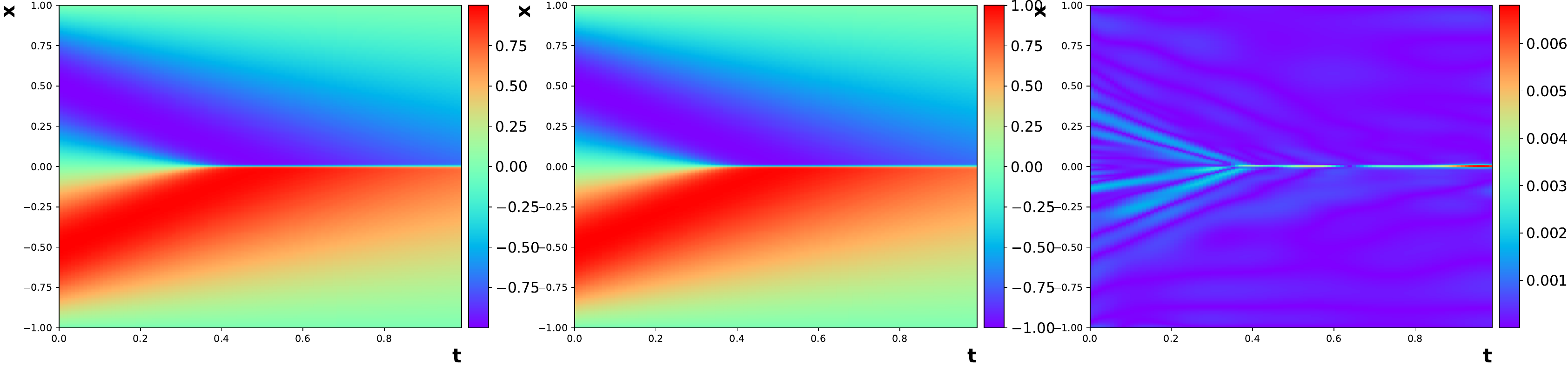}
\end{minipage}
}\\
\subfigure{
\begin{minipage}{0.98\textwidth}
    \centering
    \includegraphics[width=1.00\textwidth]{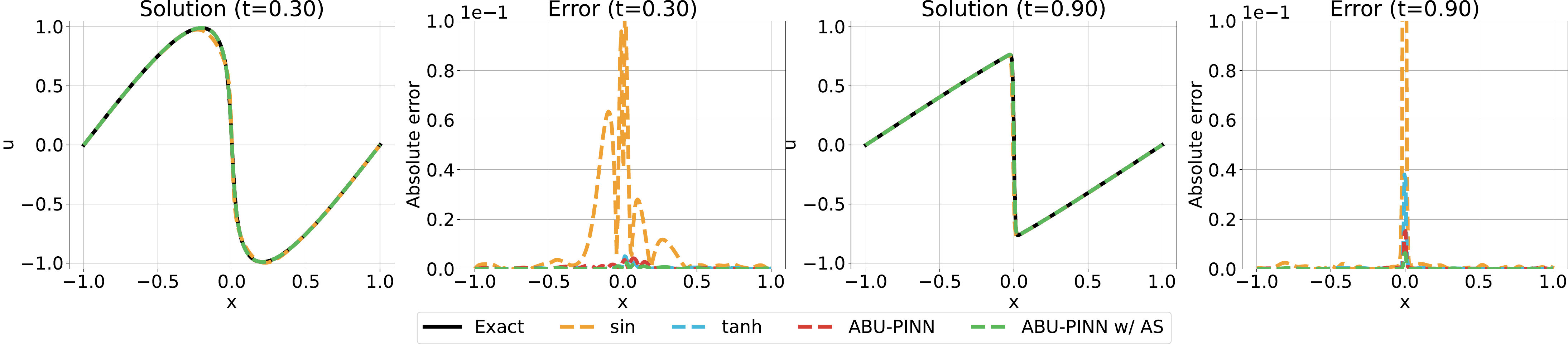}
\end{minipage}}
    \vspace{-1.3em}
    \textcolor{black}{\caption{The burgers' equation. Top: the exact solution (left), the predictions of {\name} (middle), and the absolute error between them (right). Bottom: predicted solutions at different time snapshots.}\label{fig:pred_detailed_burger}}
    
\end{figure*}

\begin{figure*}[htbp]
\subfigure{
\begin{minipage}{0.98\textwidth}
    \centering
    \includegraphics[width=1.00\textwidth]{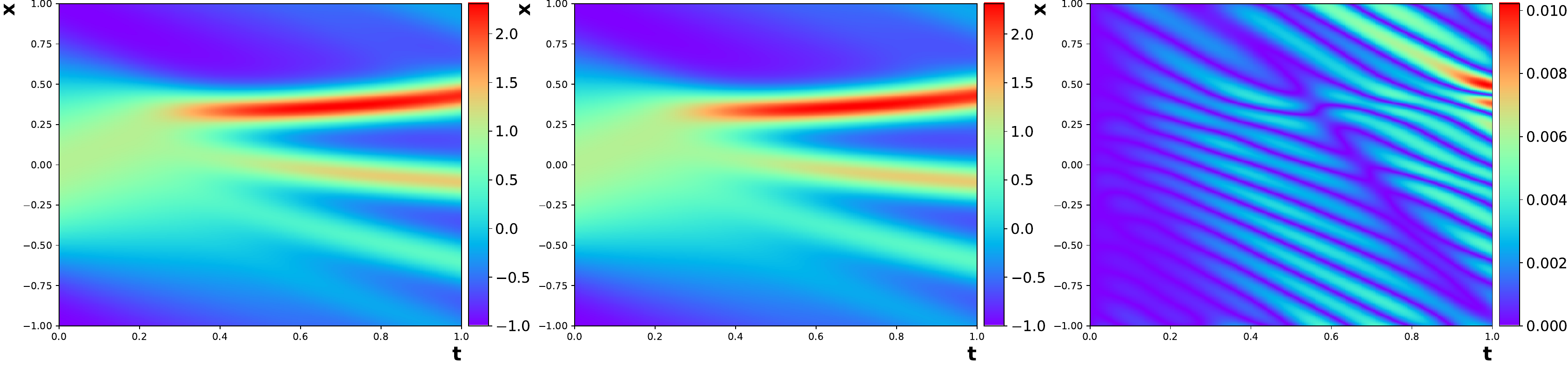}
\end{minipage}
}\\
\subfigure{
\begin{minipage}{0.98\textwidth}
    \centering
    \includegraphics[width=1.00\textwidth]{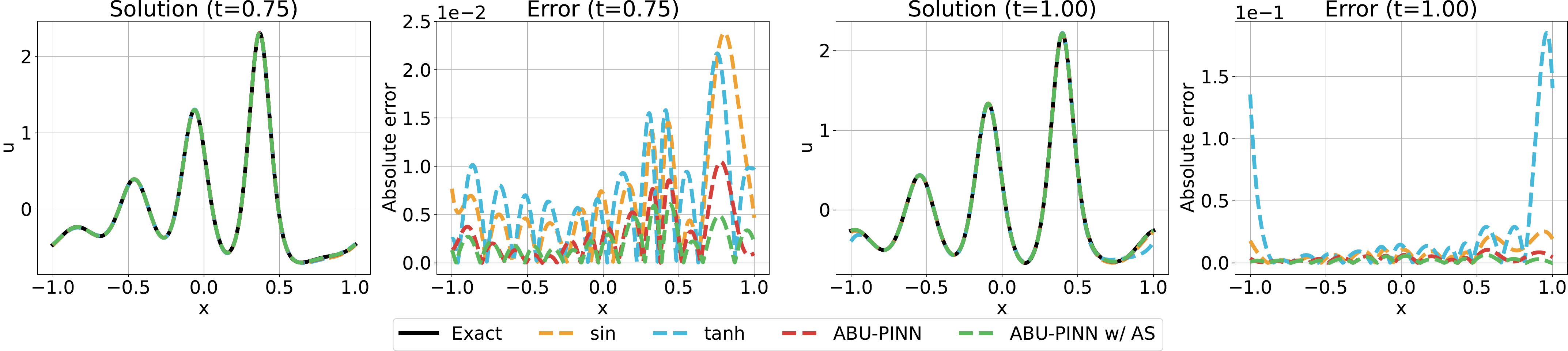}
\end{minipage}}
    \textcolor{black}{\caption{The KdV equation. Top: the exact solution (left), the predictions of {\name} (middle), and the absolute error between them (right). Bottom: predicted solutions at different time snapshots.}\label{fig:pred_detailed_kdv}}
    
\end{figure*}

\begin{figure*}[htbp]
\subfigure{
\begin{minipage}{0.98\textwidth}
    \centering
    \includegraphics[width=1.00\textwidth]{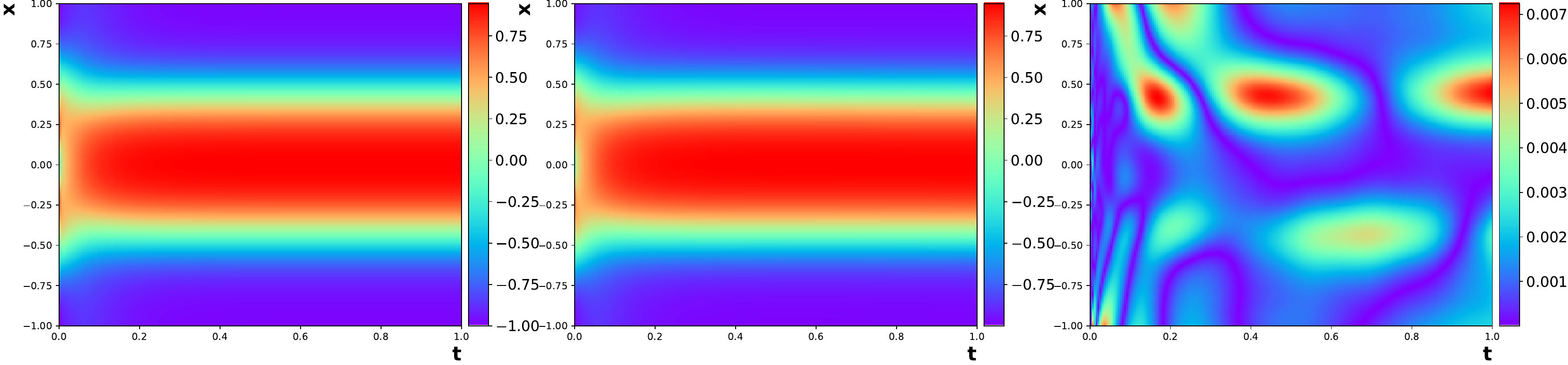}
\end{minipage}
}\\
\subfigure{
\begin{minipage}{0.98\textwidth}
    \centering
    \includegraphics[width=1.00\textwidth]{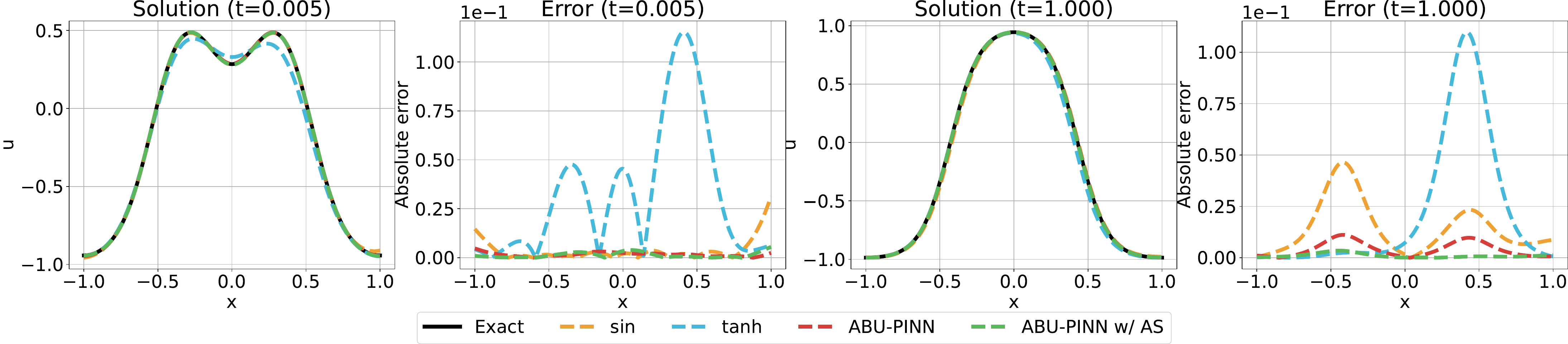}
\end{minipage}}
    \textcolor{black}{\caption{The Cahn-Hilliard equation. Top: the exact solution (left), the predictions of {\name} (middle), and the absolute error between them (right). Bottom: predicted solutions at different time snapshots.}\label{fig:pred_detailed_ch}}
\end{figure*}

\begin{figure*}[ht]
\subfigure[The Burgers' equation]{
\begin{minipage}{.49\textwidth}
    \centering
    \includegraphics[width=0.9\textwidth]{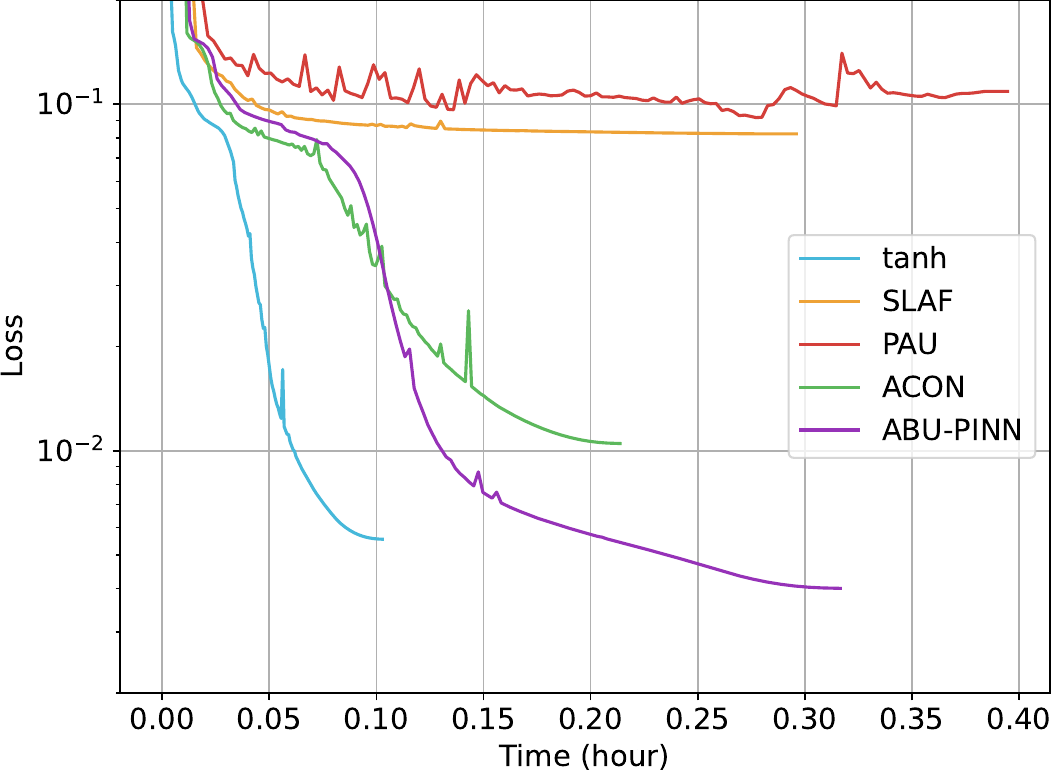}
\end{minipage}}
\subfigure[The Allen-Cahn equation]{
\begin{minipage}{.49\textwidth}
    \centering
    \includegraphics[width=0.9\textwidth]{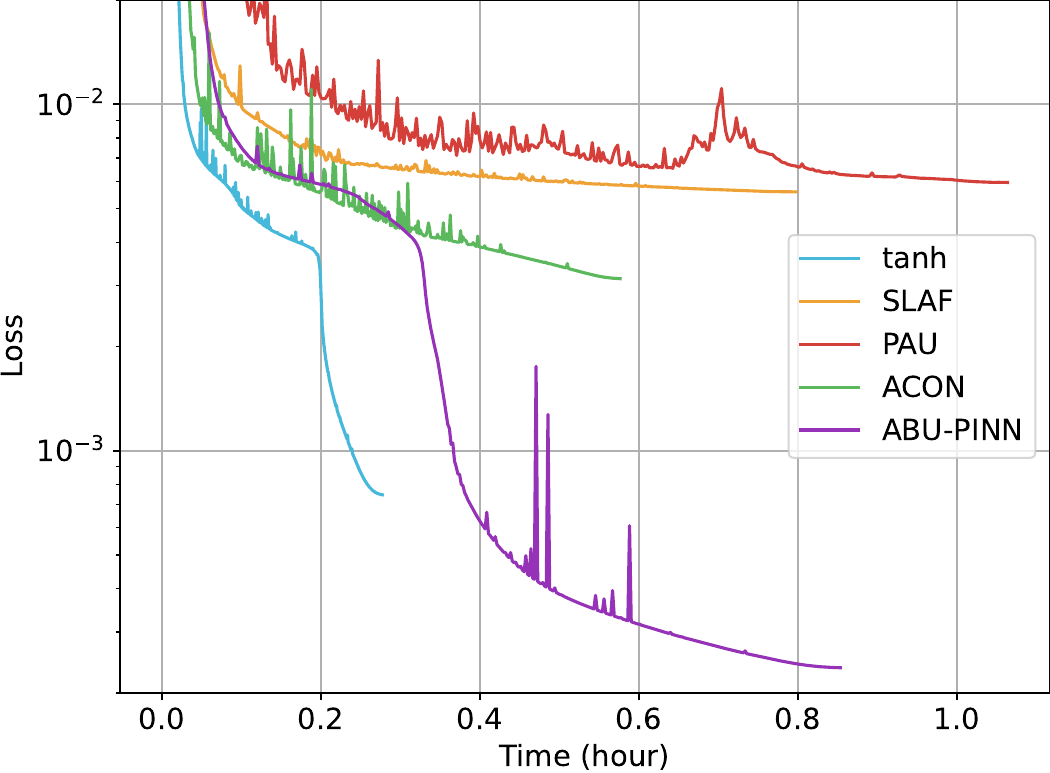}
\end{minipage}}
    \textcolor{black}{\caption{The training losses of different activation functions against training time (hour).}\label{fig:training_loss_time_log}}
\end{figure*}

\begin{figure*}[htbp]
\subfigure[$\mathrm{GELU}$ ($L_2$ relative error: 4.60\%)]{
\begin{minipage}{0.95\textwidth}
    \centering
    \includegraphics[width=1.00\textwidth]{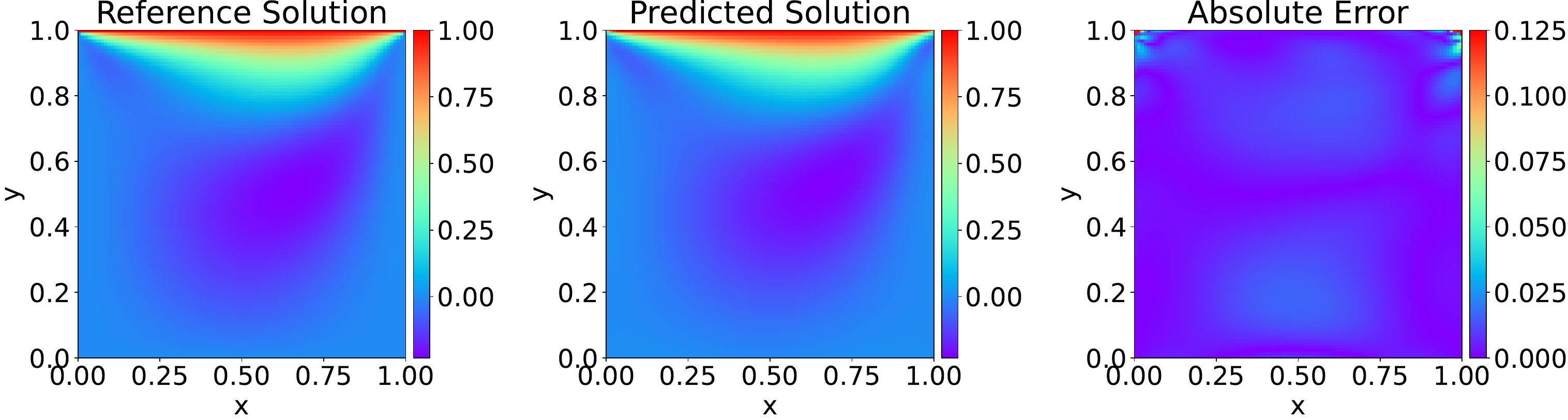}
\end{minipage}}
\subfigure[ACON ($L_2$ relative error: 4.83\%)]{
\begin{minipage}{0.95\textwidth}
    \centering
    \includegraphics[width=1.00\textwidth]{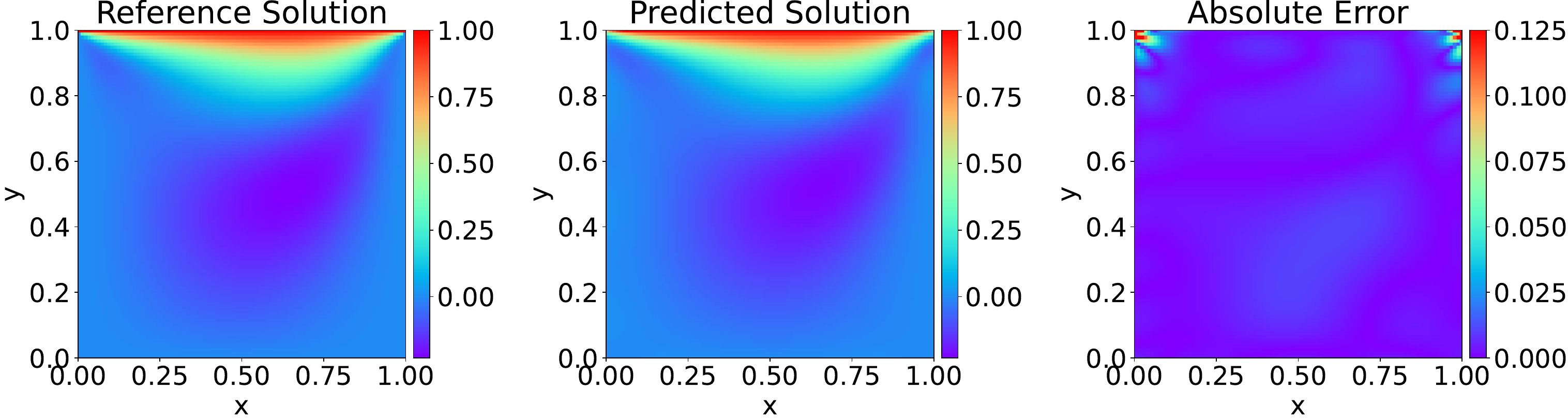}
\end{minipage}}
\subfigure[{\name} ($L_2$ relative error: 3.92\%)]{
\begin{minipage}{0.95\textwidth}
    \centering
    \includegraphics[width=1.00\textwidth]{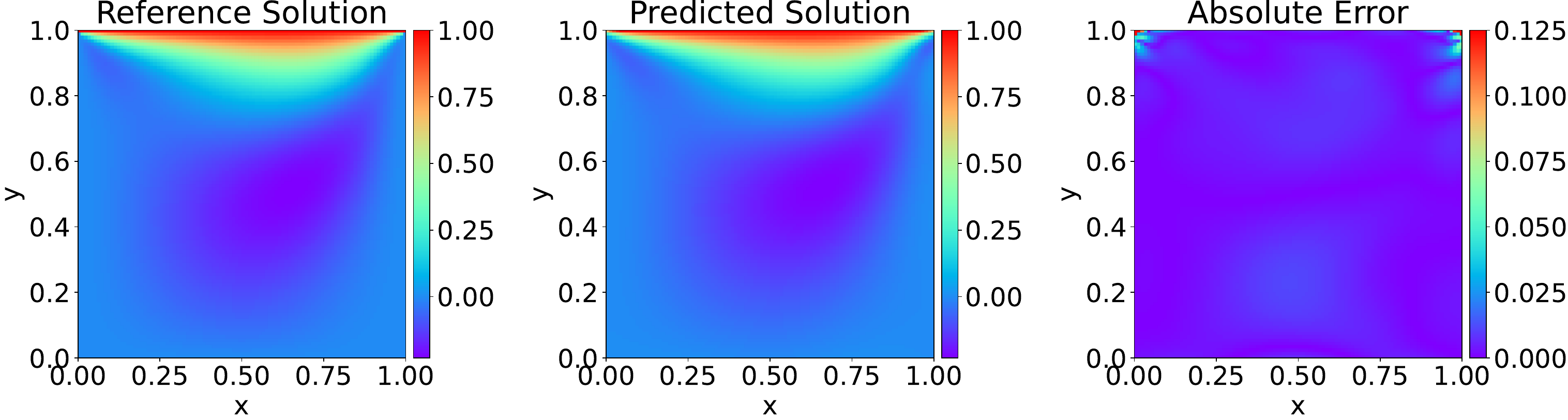}
\end{minipage}}
\subfigure[{\name} w/ AS ($L_2$ relative error: 3.50\%)]{
\begin{minipage}{0.95\textwidth}
    \centering
    \includegraphics[width=1.00\textwidth]{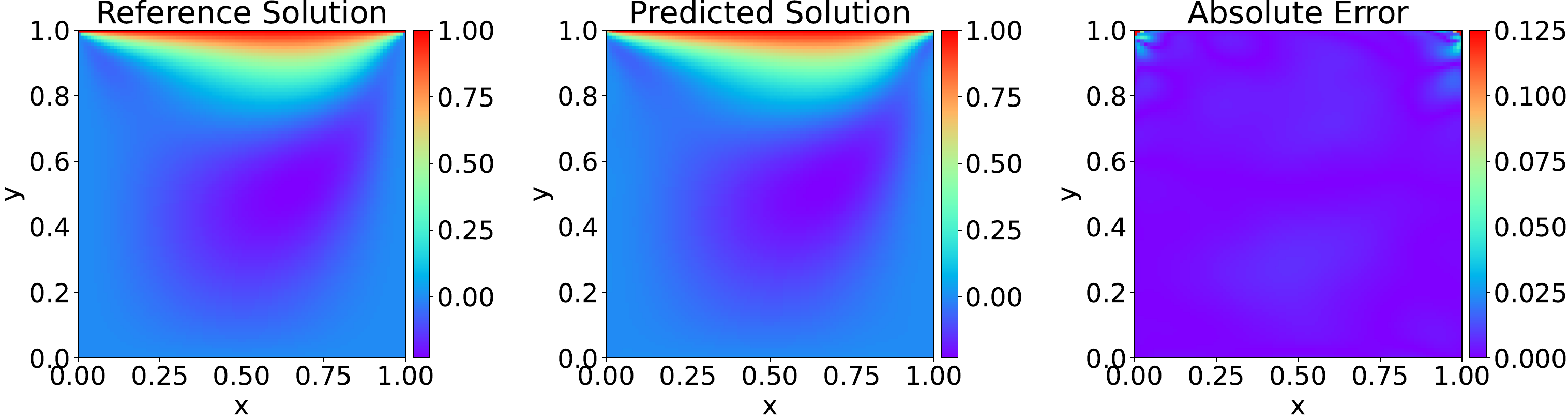}
\end{minipage}}
    \textcolor{black}{\caption{Flow in a lid-driven cavity, x-component of velocity $u$.}\label{fig:pred_detailed_flow_in_u}}
    
\end{figure*}

\begin{figure*}[htbp]
\subfigure[$\mathrm{GELU}$ ($L_2$ relative error: 0.56\%)]{
\begin{minipage}{0.95\textwidth}
    \centering
    \includegraphics[width=1.00\textwidth]{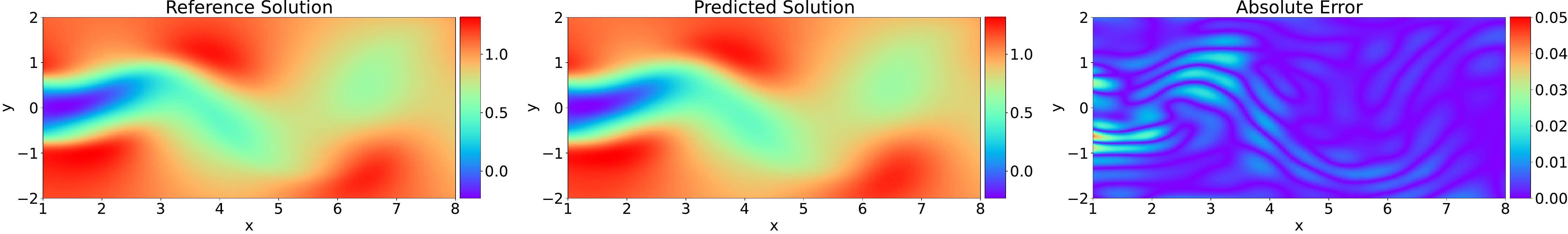}
\end{minipage}}
\subfigure[ACON ($L_2$ relative error: 0.83\%)]{
\begin{minipage}{0.95\textwidth}
    \centering
    \includegraphics[width=1.00\textwidth]{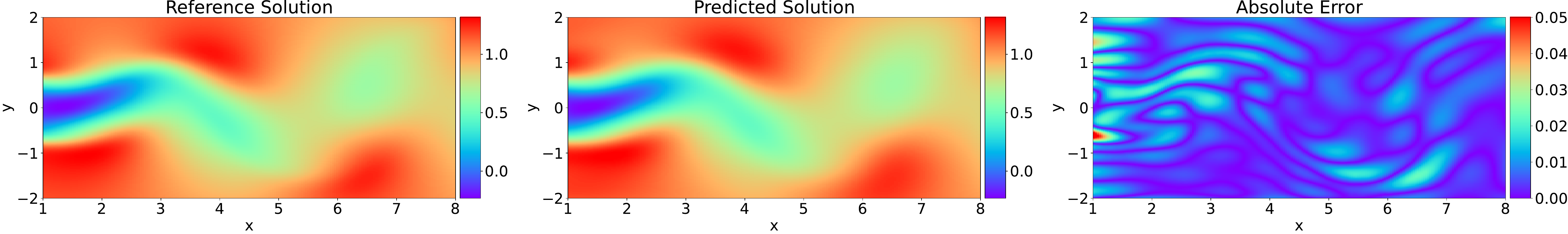}
\end{minipage}}
\subfigure[{\name} ($L_2$ relative error: 0.42\%)]{
\begin{minipage}{0.95\textwidth}
    \centering
    \includegraphics[width=1.00\textwidth]{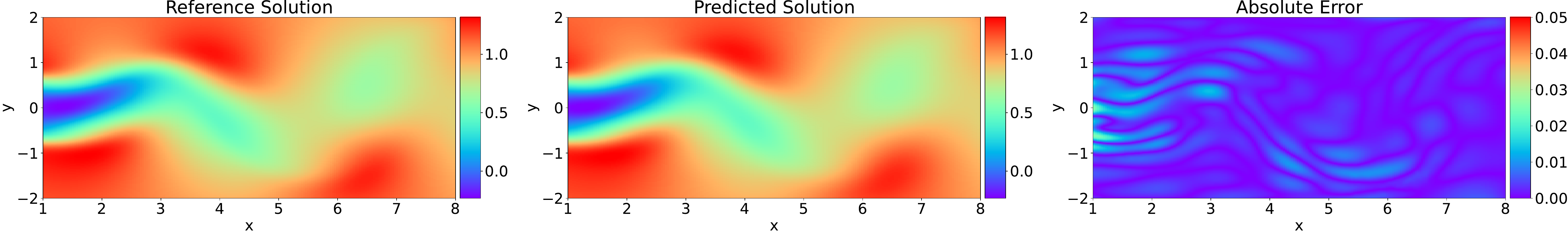}
\end{minipage}}
\subfigure[{\name} w/ AS ($L_2$ relative error: 0.28\%)]{
\begin{minipage}{0.95\textwidth}
    \centering
    \includegraphics[width=1.00\textwidth]{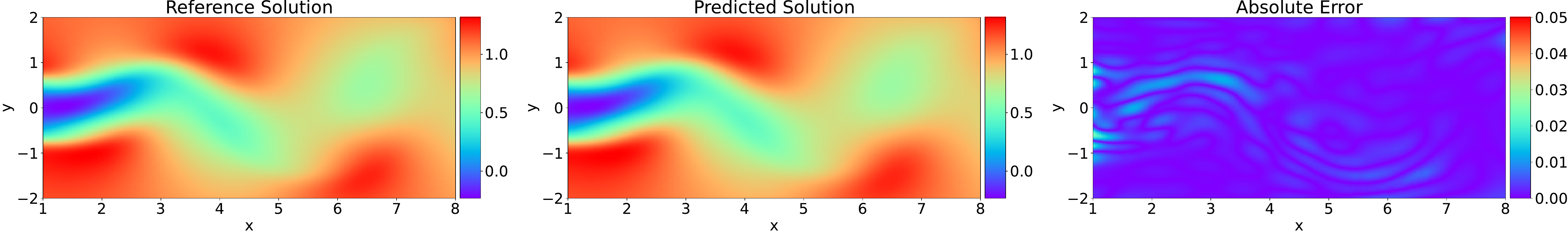}
\end{minipage}}
    \textcolor{black}{\caption{Flow past a circular cylinder, x-component of velocity $u$ (t=2.5).}\label{fig:pred_detailed_flow_past_u}}
    
\end{figure*}

\begin{figure*}[htbp]
\subfigure[$\mathrm{GELU}$ ($L_2$ relative error: 1.93\%)]{
\begin{minipage}{0.95\textwidth}
    \centering
    \includegraphics[width=1.00\textwidth]{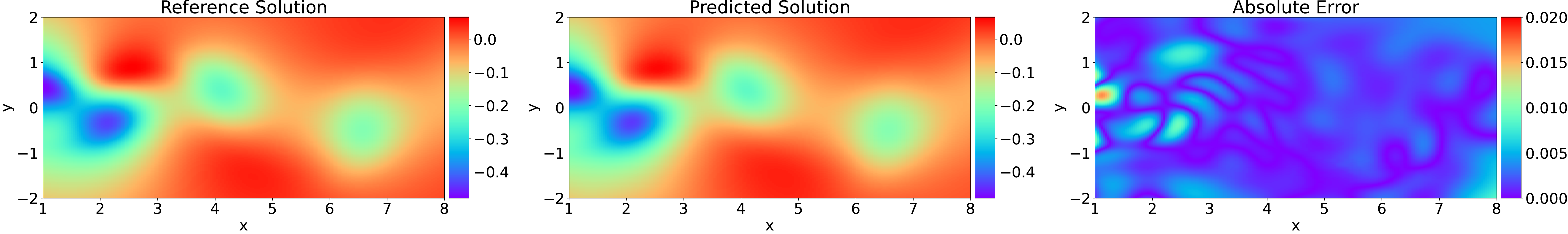}
\end{minipage}}
\subfigure[ACON ($L_2$ relative error: 3.16\%)]{
\begin{minipage}{0.95\textwidth}
    \centering
    \includegraphics[width=1.00\textwidth]{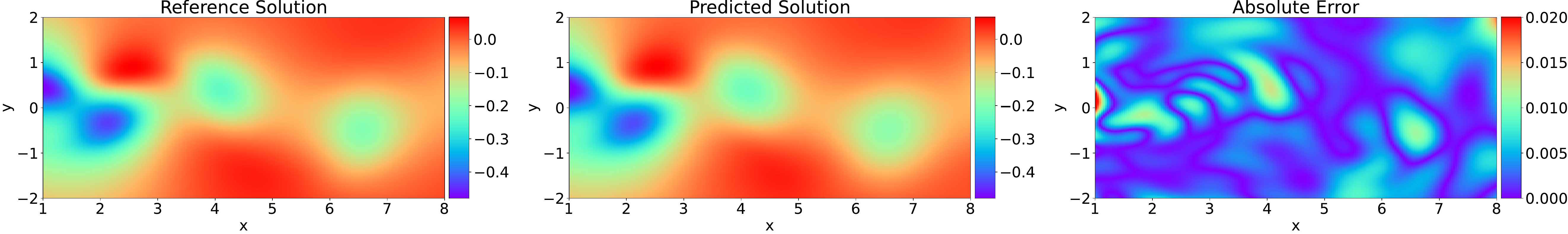}
\end{minipage}}
\subfigure[{\name} ($L_2$ relative error: 1.68\%)]{
\begin{minipage}{0.95\textwidth}
    \centering
    \includegraphics[width=1.00\textwidth]{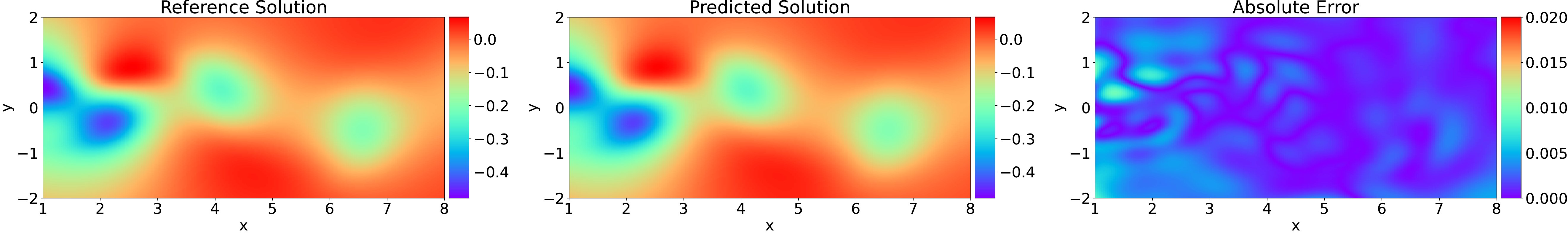}
\end{minipage}}
\subfigure[{\name} w/ AS ($L_2$ relative error: 1.56\%)]{
\begin{minipage}{0.95\textwidth}
    \centering
    \includegraphics[width=1.00\textwidth]{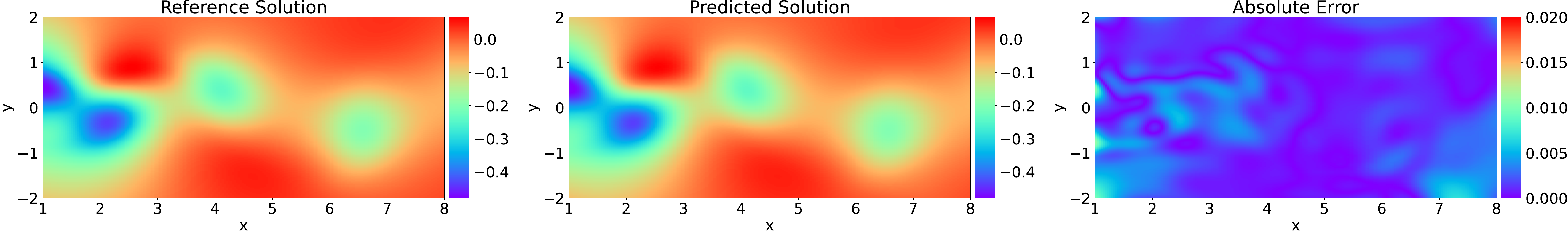}
\end{minipage}}
    \textcolor{black}{\caption{Flow past a circular cylinder, pressure $p$ (t=2.5).}\label{fig:pred_detailed_flow_past_p}}
    
\end{figure*}

\begin{figure*}[htbp]
\subfigure[$\mathrm{sin}$]{
\begin{minipage}{.47\textwidth}
    \centering
    \includegraphics[width=0.9\textwidth]{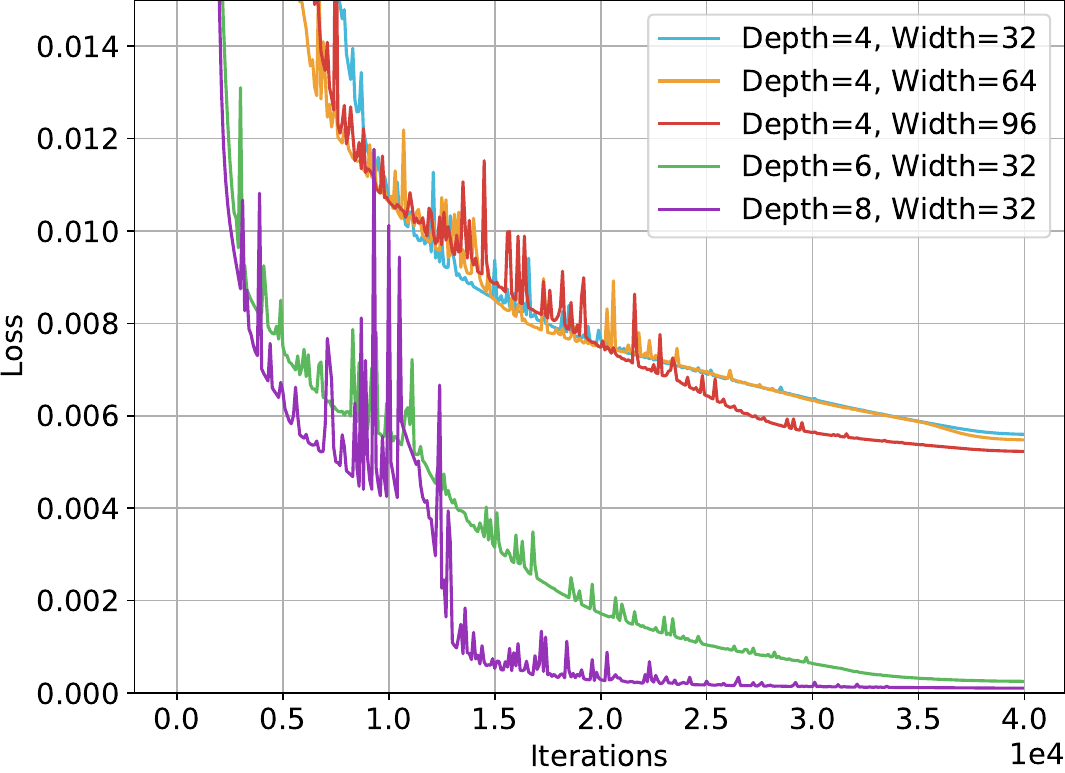}
\end{minipage}}
\subfigure[$\mathrm{tanh}$]{
\begin{minipage}{.47\textwidth}
    \centering
    \includegraphics[width=0.9\textwidth]{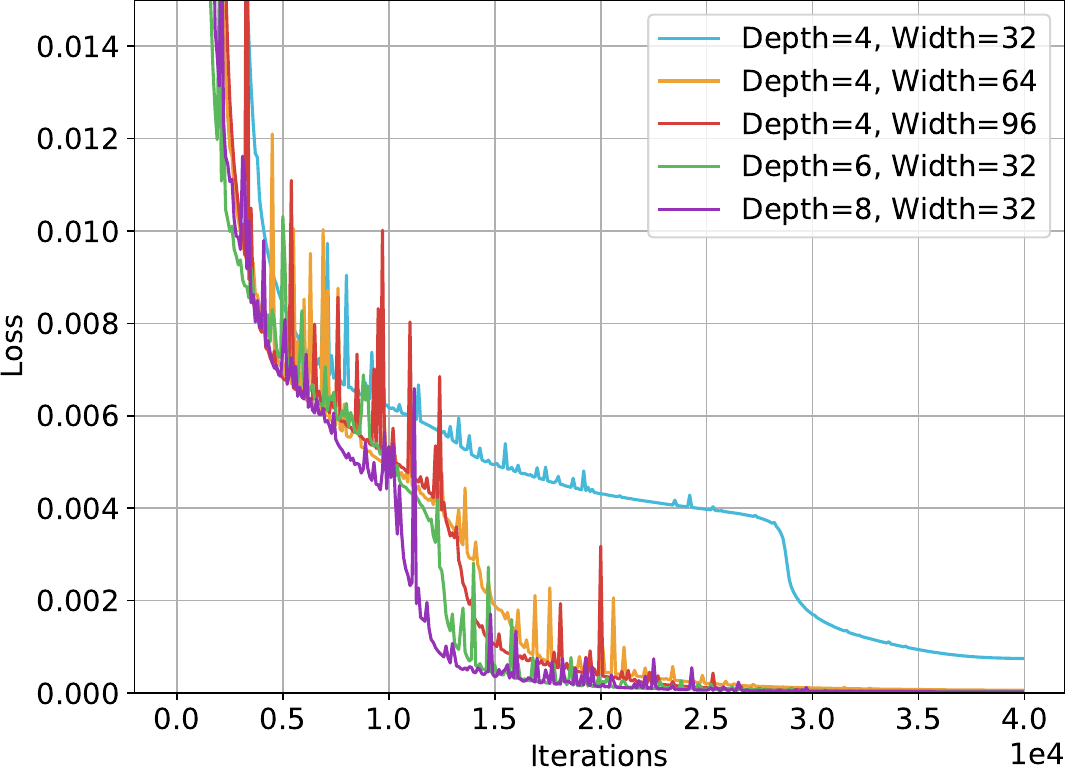}
\end{minipage}
}\\
\subfigure[SLAF]{
\begin{minipage}{.47\textwidth}
    \centering
    \includegraphics[width=0.9\textwidth]{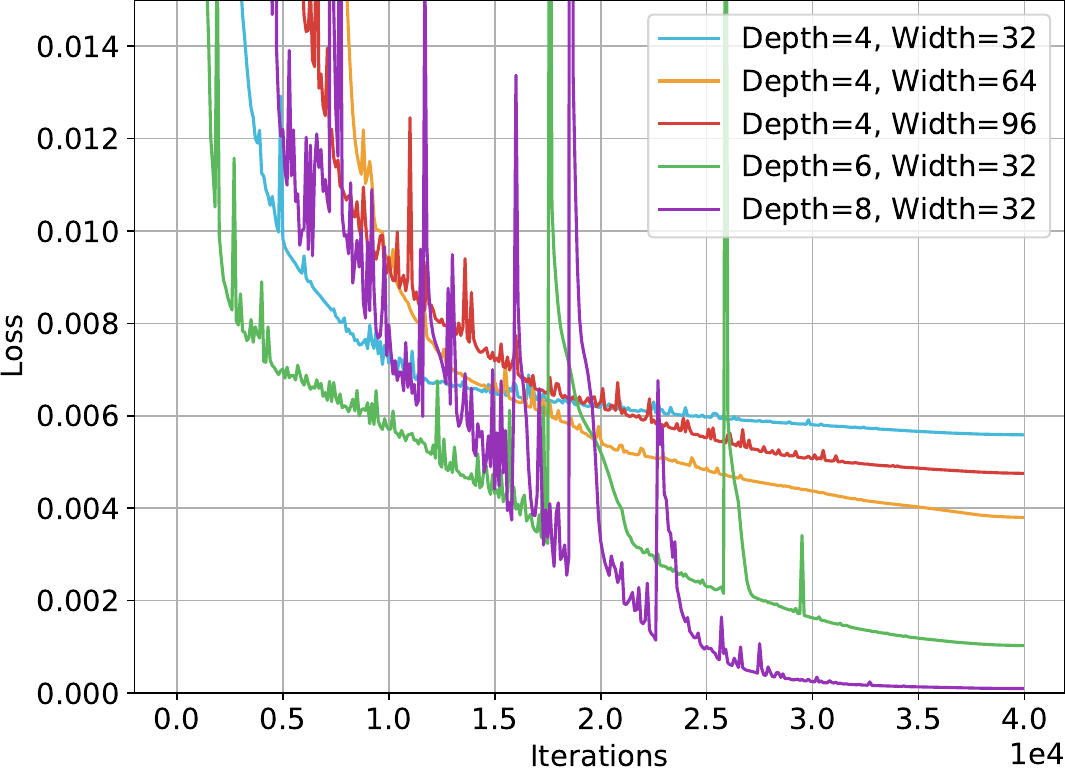}
\end{minipage}}
\subfigure[PAU]{
\begin{minipage}{.47\textwidth}
    \centering
    \includegraphics[width=0.9\textwidth]{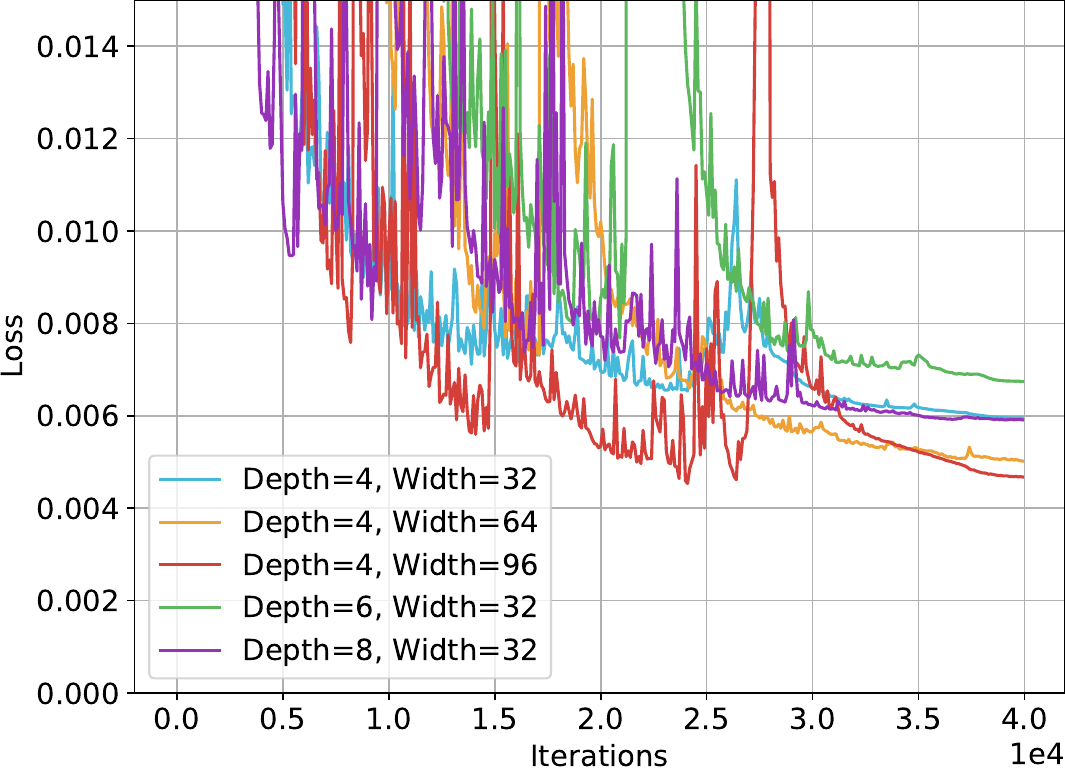}
\end{minipage}
}\\
\subfigure[ACON]{
\begin{minipage}{.47\textwidth}
    \centering
    \includegraphics[width=0.9\textwidth]{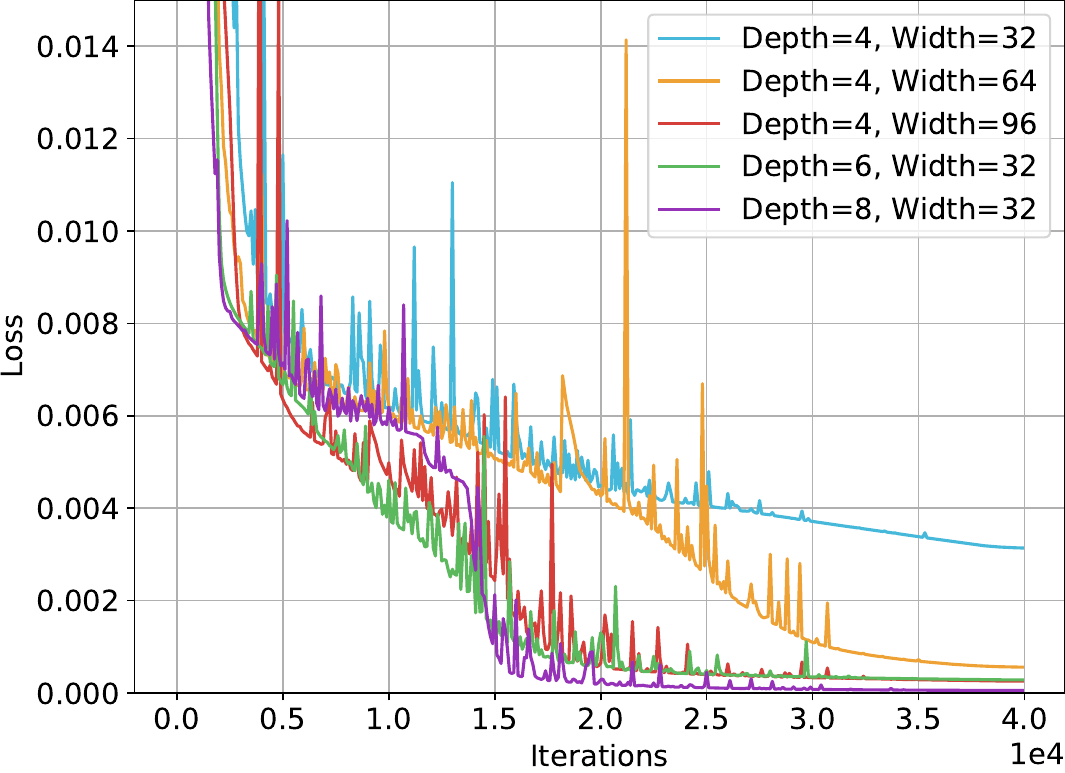}
\end{minipage}}\subfigure[\name]{
\begin{minipage}{.47\textwidth}
    \centering
    \includegraphics[width=0.9\textwidth]{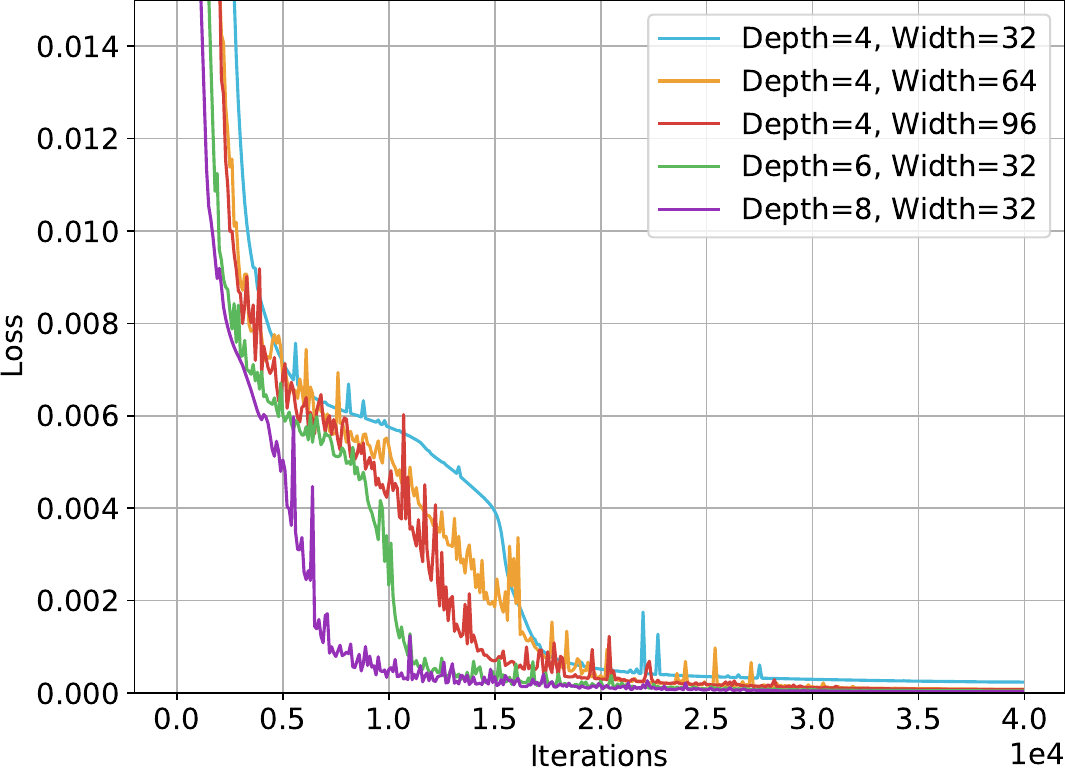}
\end{minipage}
}
    \textcolor{black}{\caption{The training losses of different activation functions across various network depths and widths.}\label{fig:train_loss_diff_network}}
    
\end{figure*}

We present the predicted solutions of {\name} for Burgers' equation in \cref{fig:pred_detailed_burger}, KdV equation in \cref{fig:pred_detailed_kdv}
, and Cahn-Hilliard equation in \cref{fig:pred_detailed_ch}. We show the training losses of different activation functions against training time in \cref{fig:training_loss_time_log}.\cref{fig:pred_detailed_flow_in_u} shows the predicted x-component of velocity $u$ in the case of flow in a lid-driven cavity. For the flow past a circular cylinder, we presents the reference and predicted solutions of  stream-wise velocity component $u$ and the pressure filed $p$ in \cref{fig:pred_detailed_flow_past_u} and \cref{fig:pred_detailed_flow_past_p}, respectively. Additionally, we provide the training curves of different activation functions across various network depths and width in \cref{fig:train_loss_diff_network}.

\bibliographystyle{elsarticle-num} 
\bibliography{reference}

\end{document}